\documentclass{article}

\usepackage{PRIMEarxiv}
\usepackage{amsmath}
\usepackage{amssymb}
\usepackage{tabularx}
\usepackage[numbers,sort&compress]{natbib}\usepackage{algorithm}
\usepackage{algpseudocode}
\usepackage{bbm} 
\usepackage[utf8]{inputenc} 
\usepackage[T1]{fontenc}    
\usepackage{hyperref}       
\usepackage{url}            
\usepackage{booktabs}       
\usepackage{amsfonts}       
\usepackage{nicefrac}       
\usepackage{microtype}      
\usepackage{lipsum}
\usepackage{fancyhdr}       
\usepackage{graphicx}       
\graphicspath{{media/}}     

\title{Hybrid Semantic Context-Enhanced Ensemble Learning for Wind Power Ramp-Event Forecasting and Uncertainty-Aware Evaluation
 
}

\author{
  Momina Liaqat Ali \\
  Department of Industrial and Systems \\ Engineering,
  University of Tennessee, Knoxville \\
  Knoxville, TN 37996, USA. \\
  \texttt{mali35@vols.utk.edu} \\
  \And
  Muhammad Abid \\
  Department of Mechanical and Aerospace \\ Engineering,
  University of Tennessee, Knoxville \\
   Knoxville, TN 37996, USA. \\
  \texttt{mabid@vols.utk.edu} \\
  \And
  Muhammad Abdullah \\
  Department of Mechanical Engineering, \\
  Pakistan Institute of Engineering and \\ Applied Sciences, 
  Islamabad, Pakistan. \\
  \texttt{msme2313@pieas.edu.pk} \\
  \And
  Aneela Zameer\thanks{Corresponding author. Email: aneelaz@pieas.edu.pk} \\
  Department of Computer Science, \\
  Pakistan Institute of Engineering and \\ Applied Sciences, 
  Islamabad, Pakistan. \\
  \texttt{aneelaz@pieas.edu.pk} \\
}

\begin{document}
\maketitle

\begin{abstract}
Wind power ramp events which are sudden, large swings in turbine output over short windows are difficult to estimate, and standard models often miss them. Hybrid forecasting approach is built which augments semantic context to ramp-event forecast. Rather than applying an extensive language model directly to predict turbine operating data, we have implemented a pipeline where turbine operating data is converted to simplified text, which is then converted to dense embeddings to be used as inputs for ensemble models incorporated with other features. Testing runs are performed at multiple intervals within the SDWPF dataset, including 10-minute, 30-minute, and 60-minute horizons, with ramp events constituting the highest change in future power output. We check robustness against autoregressive, LSTM, and GRU baselines plus several ensemble configurations, using Diebold-Mariano tests and bootstrap confidence intervals, and we vary the ramp threshold, compress the embeddings with PCA, and validate externally on Kaggle SCADA and NREL data with uncertainty-aware scoring. The semantic-context features produce negligible yet statistically significant gains over the baselines in multiple paired ensemble runs, most clearly at the 30- and 60-minute horizons where these gains hold across different ramp-threshold definitions, and PCA compression helps in some longer-horizon cases. The best context-augmented ensembles rank near the top overall, though the GRU model still posts the lowest ramp-event RMSE at 30 and 60 minutes. External tests confirm the error reduction generalizes across datasets, but the size of the gain depends on both model and dataset. Prediction intervals cover most test cases well but weaken during ramp events, pointing to a localized shift in the data distribution.
\end{abstract}


\section{Introduction}
\label{sec:introduction}

The inclination of wind energy into contemporary electricity systems displays an urge on the need for wind power forecasting mainly for the ease of running of the grid, energy timetabling, allocation of a backup supply of electricity in case of shortage, the ability to compete in electricity markets, or integrating renewables and others. This is mainly because wind energy being a weather dependent source, and loads of other additional variables like how a turbine operates, how the atmosphere twists in that area amongst others makes it very unlike conventional generation. There are peculiar changes in power output that can occur very fast over a short period of time. Wind power ramp events are one of such changes and to a lesser or larger extent such events contribute to the load variations thus increasing the level of reserves in the system and giving headaches for system operators because of changes in power generation resulting in the deviation to be managed.

Several methods have been suggested to address the issue of wind power forecasting based on the physical, statistical and time series techniques. The physical models resort to the nwp and turbine characteristics, the statistical and autoregressive models on the other hand are founded on the previous power generation and the empirical observations on meteorological variables to tackle the granger causality tests. The archaic ways of forecasting wind power using techniques like persistence, autoregressive models, ARIMA-type models and lasso represent benchmarks, very especially for very short time horizons. Nevertheless, they neglect the challenges of representing nonlinear turbine behaviors and ramp dynamics.

With the issue of short-term wind power forecasting, machine learning and deep learning models have come to the rescue. Tree-based ensemble methods like Random Forest, Extra Trees, Gradient Boosting, XGBoost, LightGBM, and CatBoost can capture the non-linear relationship between the lagged power, wind speed, wind direction, the states of the turbines, temporal variables, and spatial information. Another category is that of deep sequence models, which consist of long short-term memory or LSTM, and gated recurrent unit or GRU networks, which can capture the dependencies between entities in the sequence that are affected. Despite such GPU accelerated advancements, quite a good number of the prevailing forecasting models run on extreme of numerical functions/minimal use of manually engineered features, aspects which do not exhaustively enumerate regimes in which the transition from lows to high speeds in wind and the case of the turbine producing less power than it should is observed.

Recent advancements in large language models and language-based learning of the internal structure of the language have opened up new opportunities and the prospect of using semantic representations for the purpose of time-series modeling. However, there is no guarantee that LLMs or any such modelling technique will always beat the traditional or ad-hoc forecasting models. It is all depended on how effectively memory about the series is saved and or inferred, integrated that into factor and used it for forecasting purposes. In the present research, it is not about the comfort of the automatic usage of an LLM as a straight white box forecaster. Instead, in the context of the study, the black-box forecasting capability of an LLM known as an automatic encoder is renounced, and LLM-driven semantic embeddings are used as auxiliary operational-context features. For example, the numerical information regarding turbines’ and weather information includes the evaluation of the speed and wattage, focused on the wind direction including the status of whether the power has been increased or decreased recently, its unforeseeable change, hidden change, and the risk of changing suddenly over a short time period.

This work introduces a new method of wind ramp-event forecasting based on semi-structured predictive modeling in wind energy, which refers to the class of strategies that use numerical predictors, symbolic environmental context types, textual descriptions, and even more complicated models that use the representations that come from the LLM and other rich semantic annotation strategies. The main point of reference here is the ramp events in wind power production within the SDWPF wind power data sets applied for the horizons of 10 minutes, 30 minutes, 60 minutes respectively. The analysis is focused on wind power shifts of various event magnitudes, with an additional test being conducted on ramp-threshold sensitivity analysis in terms of the top 10\%, 15\%, 20\% ramps. The encouraging technique includes cross models implemented (simple, autoregression, probabilistic ensemble methods, boosting, recurrent neural networks) aspects such as, simple cross models, autoregression, probabilistic ensemble methods, boosting, recurrent neural networks, and context-aware boosting. Finally, the most specialised ones; PCA-based semantic embeddings, repeated scaling of the model returns, and univariate conditional value at risk models with median estimation without control.

A comprehensive empirical examination of the issue includes a relevant counter-factual approach with no perturbation ('no LLM') comparison, where a specific model is built with and without its rich off-the-shelf LLM-derived extensions, then it examines how much more useful the semantic operational context is irrespective of the learning algorithm used. Participating in an ensemble approach at several settings and time points, the empirical gains are minimal but not entirely trivial. Statistical tests confirm some value addition, especially the 30- and 60 training hours. Bootstrap confidence intervals together with Diebold-Mariano test is applied for assessing whether these found improvements are greater than variations tied to sampling. The so-called model competition also contains deep neural networks models thus in the competition, and the favorable results were achieved by GRU with the most economic forecasting of the ramp event errors for the largest horizons, while the LLM ensemble solutions could reach similar results and introduce some interpretable additional context.

Furthermore, in order to assess whether the developed models are useful in practice, their external validation was conducted against the SCADA data so as to determine other factors that could be outside the factor. A research concluded that constructing of models with semantic and operational context predictions; and then using them made through the LLM should provide better accuracy in predicting models across all the data sets when compared to other defense algorithms as this was the goal itself of the research. Consequently, SDWPF is seen to be the main testbed for focused ramp-event analysis so that the comparative strength of the models may be gauged, while the concerned reader may be interested in the wider applicability of the method concerning sustained forecast generation.

\begin{figure}[htbp]
    \centering
    \includegraphics[width=1.0\textwidth]{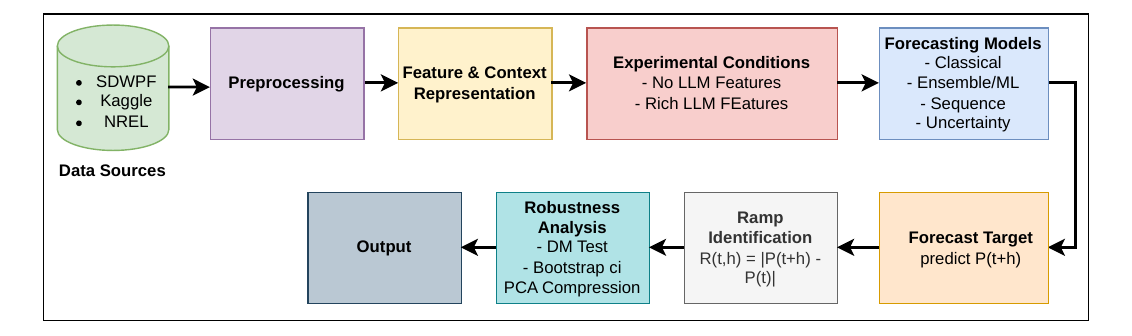}
    \caption{Detailed workflow of the suggested combined semantic-contextual based wind power estimation system. The design covers data preparation, construction of numerical/semantic contextual content, modeling enhancements, verification of statistical measures, ramping sensitivity study, external proof of the model, and calculation of uncertainties.}
    \label{fig:overall_workflow}
\end{figure}

Figure~\ref{fig:overall_workflow} illustrates the proposed workflow. There are a number of stages at which the wind turbine data is processed. In stage I, the data is subjected to a number of operations: handling of missing values, handling of invalid powers, partitioning by sort order, and reshaping based on features. Stage II makes use of numeric lagged and rolling features, symbolic context-operator categories, text-based representations, rich LLM semantic embeddings, and PCA-compressed derivations of embeddings. At the following stage, stage three, it is possible to build traditional, ensemble, sequential models devoted to the forecasting task, and models receptive to the existence of uncertainty. These models are being tested under the following environments: full-test condition, ramp-event situation, statistical robustness scenario, and threshold sensitivity constraints.

The main contributions of this study are as follows:

\begin{enumerate}
    \item Rapidity of ramp-events is forecasted using a novel semantic operational-context representation that is created by waster-of-wind turbine operational states to more easily comprehended and information-packed LLM-derived representations.

    \item In this work, we create a paired design comparing a rich LLM with no LLM, to examine the value of semantics, while keeping the same forecasting family of models.

    \item  In this paper, a direct comparison is made between the plain-persistence, the autoregressive linear base-line, the ensemble models, the models based on boosting, the sequence model based on long short-term memory (LSTM), as well as gated recurrent unit (GRU) and the quantile-based models which is known as the uncertainty-masking model, GRU shows good performance in long-term forecast.

    \item The validity of the application of their estimates is checked medically using the Diebold  and Mariano strength tests and bootstrap confidence intervals as well as ramp severity analyses to forecast the ramp every group whereby their performance is within the upper 10\%, 15\% and 20\% band during the ramp.

    \item We also conduct experiments to see whether PCA-based embedding dimensional reduction can remove noise in higher dimensions of semantic embedding and thus improve selected predictions regarding wind ramp events with long horizons.

    \item The external validation checks on Kaggle SCADA and NREL wind data sets are done to further see if the enriched semantic model from LLM travels outside of the main SDWPF associated performance regimes.

    \item An assessment framework for diagnostic studies that is both sensitive to ramp-event definitions and considers updating processes in order to distinguish issues in competency with ramp-event settings versus issues in general.
\end{enumerate}

The remainder of this paper is organized as follows. Section~\ref{sec:related_work} reviews related work. Section~\ref{sec:data_problem} describes the datasets and forecasting formulation. Section~\ref{sec:methodology} presents the proposed framework. Section~\ref{sec:experiments} describes the experimental setup. Section~\ref{sec:results} presents the results and discussion. Section~\ref{sec:conclusion} concludes the paper.

\section{Related Work}
\label{sec:related_work}

\subsection{Recent Advances in Wind Power Forecasting}

The prediction of wind energy yields is one of the important issues for research because it is a phenomenon which is actually stochastic, non-linear, and subject to external environments and the setting of the wind turbine filtering element states. The most recent evaluations suggest that change has occurred such that the focus has embraced machine learning type methodologies in addition to traditional methodologies, including but not limiting to physical or statistical models, and their probabilistic versions and hybrids forecasting format \citep{yang2024surveywindml,xie2023overviewwind,wu2022comprehensivewinddl,bazionis2022probabilisticwind}. Grid planning requires accurate estimates of wind power as it has potential restoring the green house gases, energy from renewable sources, market for power and power system security of supply.

Some of the elements that influence the prediction performance are model complexity, temporal dependence, input representation, geographical variation or otherwise meteorological dependence, and model architecture \citep{yang2024surveywindml,xie2023overviewwind,energy2024statictransformerwind,han2022blslstmwind}. Moreover, short term prediction may also have good forecasts if the late wind and power measurements where there are no prior power and wind information without extreme adjustments to the model. On the other hand, the longer prediction periods mostly necessitate designs which can take advantage of enhanced temporal relationships and that the power production scheduling can be altered repeatedly, across the entire prediction period. Consequently, more recent research into wind power prediction incorporates historical power data, weather variables, lag and successor positions, time of day and year, and makes use of both traditional summary techniques as well as structured input representations.

Wind power forecasting has increasingly relied on machine learning and deep learning. Researchers have used ensemble approaches, recurrent neural networks, convolution-based models, hybrid setups, and attention-style designs to learn nonlinear links between turbine readings and weather inputs \citep{yang2024surveywindml,wu2022comprehensivewinddl,energy2024statictransformerwind,zhao2024dewp}. Earlier works demonstrated it is possible to hybridize neural and evolutionary methods for learning. In informing us about the application of such an approach, Zameer et al (2017) proposed a genetic-programming ensemble of neural networks for predicting the short-term wind speed. Qureshi’s (2017) group utilized deep neural networks with meta-regression and transfer learning to achieve the same goal as before \citep{zameer2017intelligent,qureshi2017wind}.

In the work of Shahid and colleagues, a WaveNet-LSTM approach was introduced, along with a genetic LSTM. Based on the findings mentioned in their research, the combined approach of convolutional sequence modeling with recurrent networks leads to superior performance than recurrent model alone, among other findings. Moreover, evolutionary optimization was another technique used in the development process Shahid et al. \citep{shahid2020novel,shahid2021novel}. There have been more discoveries afterwards. In particular, a GRU model is held up for weather forecasting a couple of hours from now depending on the entire size of a state, while another direction introduced additional lines, namely hybrid metaheuristic based feature selections, for enhancing the short term-ahead predictions of wind power \citep{farah2022short,mudassar2026gwo}.

In prior works in the study of time-series forecasting, it has been established that models like Transformer-based models, such as Temporal Fusion Tranformers, Informer, and PatchTSB Exploit, can be used satisfactorily for difficult patterns in sequence \citep{lim2021temporal,zhou2021informer,nie2023patchtst}. These models operate on the basis of attention and learned representations, which alleviates the problem of long and complex dependencies. However, all these models are universal and not specifically designed for wind power. Still, their results have helped drive more interest in newer sequence models for energy forecasting. A few review papers also point out something useful: there is rarely a single model type that wins on every dataset or at every forecast length. So, it makes sense to test wind-related forecasting against a mix of baselines. That mix should include classical methods, ensemble methods, and deep-learning models.

\subsection{Wind Power Ramp-Event Forecasting}

Wind power ramp events are big swings in how much power is produced, and they happen over a short span of time. Grid operators care about this since quick ups and downs in wind output can raise the need for balancing, make reserve plans harder to schedule, and add more uncertainty to everyday system control. Forecasting ramp events is tough because they show up less often than normal operating periods. Also, when they do occur, the size, how long they last, and exactly when they happen can vary a lot \citep{xie2023overviewwind,dhiman2022waveletramp,li2023transformerramp}.

The research on wind power ramp forecasting has followed several different approaches. Literature typically refers to techniques such as regression, classification, probability, attention, gcn-like, and multi-task paradigms, see, e.g. \citep{dhiman2022waveletramp,li2023transformerramp,wang2024spatiotemporalramp,peng2024graphattentionramp}. Another common aspect is whether the ramp is considered with the estimation of wind power being the first step by some existing pipelines. After that, they infer ramp behavior from the future curve. Other lines of work skip the intermediate power forecast and instead predict the ramp event itself. This includes things like whether a ramp happens, its group or label, its size, and other related attributes. More recent efforts add extra angles. Researchers have used ramp rate features, built probabilistic views of ramp behavior, and tried weather guided segmentation schemes. There are also risk centered formulations and ways to augment data for uncommon ramp examples \citep{sun2024varramp,yang2024rampsegmentation,frontiers2022spatiotemporalramp,visualization2023rampdetection}.

One key issue is that ramp events make up only a small part of the full operating distribution. So, good results on the overall test set do not always mean the model does well when ramps happen. For that reason, earlier work suggests checking ramp behavior on its own, apart from normal forecasting cases \citep{dhiman2022waveletramp,li2023transformerramp,wang2024spatiotemporalramp}. Here, we follow that idea. We measure forecasting accuracy on the whole test set as well as on the ramp-event subsets. We define consecutive ramp events as those with the highest dissimilar megawatt changes. Under this context, the sensitivity analysis assesses to green zone all ramp scenarios with such changes at the 10\%, 15\%, and 20\% quantiles levels. This ensures that the ouput is not affected by one ramp value alone.

\subsection{Feature Representation and Context-Aware Forecasting}

Effective feature engineering is critical to enhancing the accuracy of a wind power forecast since a forecast is not solely reliant on the performance of an algorithm but also the description of the states of the turbine and the weather. A few of the usual parts include recent active power, wind velocity and direction, and some other meteorological factors along with time lapse features and those incorporating an extension in time, from previous observations to recent ones including a noisy signal \citep{yang2024surveywindml,xie2023overviewwind,wu2022comprehensivewinddl}. Considering the aforementioned inputs will enable the model to proof how the last wind turbine behaved over some period, the prevailing short term wind evolution given the forecast hour and how the ambient condition changes.

Newer approaches have a tendency to pertain to more than just the raw signals and process them in the right manner. Besides taking raw signals and processing them, an approach also includes feature extraction and learned representations. These involve hybrid structures, including the addition of static context and increases challenging deep nonlinear mappings \citep{energy2024statictransformerwind,han2022blslstmwind,zhao2024dewp}. The hypothesis is that hyper parameters of the model that defines the structure of brain connectivity may contain valuable information about the variables as well. These patterns may not show up when only single measurements are used on their own.

This raises another question that this research is focused upon addressing; whether simple terms like ‘how’s the turbine’ can further predict its performance than, for example, numerical data. In some cases certain numerical patterns may fit into specific operational context. Such contexts can range from a rapid increase in wind speed, non-steady power, changes in wind directions, insufficient turbine output or frequency of steep ramps. Rather than proceed to derive that language-oriented effects will be more effective, this study seeks to clarify this issue through direct experiments. Specifically, the author evaluates the effect of verbal cues and the claim features extracted from natural language in aiding forecasting in a measurable manner.

The framework then considers several different conditions for encoding the turbine context. This may include straightforward numeric variables that describe the turbine, depending on the simple to advanced nature of the turbine, to using operating states which involve processing any categories or hulligrams directly through the use of TF-IDF features made from text summaries. This study also tests for embedding, or representations of symbols, which are dense embeddings drawn from a large language model. This involves the use of embeddings as well as their truncated linear combinations because of PCA. Altogether, the comparison remains valid because it questions whether there is any enhancement to the predictive capability of lagged and weather potentials by including descriptions of a state of a turbine in the form of text.

\subsection{Transformer and Foundation-Model Approaches for Time-Series Forecasting}

It is worth noting that currently, Transformer models used for time series data prediction have rapidly increased. There were certain concepts discussed in Informer that include self-attention tricks that are helpful for working with long dependency sequences. More recently, Autoformer also inherits signal decomposition plus a truncated repeat module for long horizon tasks \citep{zhou2021informer,wu2021autoformer}. A novel dimension of the concet of decomposition has been uncovered by FEDformer. Whereas, PatchTST extended the domain altering representation, dealing with ‘patch’ domain, which works well also for multivariate time series \citep{zhou2022fedformer,nie2023patchtst}.

Later models also moved in different directions. TimesNet looks at how time changes. iTransformer uses an inverted attention layout. TimeXer brings in external variables. TimeMixer focuses on multiscale temporal pieces \citep{wu2023timesnet,liu2024itransformer,wang2024timexer,liu2024timemixer}. Wind power forecasting fits these ideas for a reason. Turbine output varies over time and it is not linear. It also depends on what happens in the past, plus weather factors that come from outside the turbine. Even so, newer Transformer designs do not always beat simpler methods. This can change with the dataset, the forecast length, and how you score results. So it matters to test context-enhanced forecasting against both classic machine-learning baselines and other dedicated sequence models. Relying on only one model type is risky.

\subsection{LLMs and Foundation Models for Time-Series Forecasting}

Large language models and foundation models have been used for time-series prediction in a few ways. One line of work, called Time-LLM, changes the input time data into a form that lets a pretrained language model do forecasting. AutoTimes takes a different path and treats a large language model as an autoregressive forecaster \citep{jin2024timellm,liu2024autotimes}. Other papers look at zero-shot prediction. They also study how pretrained language models can be adjusted for general time-series tasks \citep{gruver2023zeroshot,zhou2023onefitsall}.

Meanwhile, foundation-style models for time series have become more common. TimeGPT, Lag-Llama, Chronos, TimesFM, and Moirai all focus on big pretrained predictors. They aim to move knowledge across datasets. Some also support zero-shot use and probabilistic outputs \citep{garza2023timegpt,rasul2024lagllama,ansari2024chronos,das2024timesfm,woo2024moirai}. These approaches are not made only for wind-power forecasting. Still, they fit a wider shift. Researchers are leaning toward pretrained and reusable representations for predicting over time.

Some recent studies also ask a practical question that if the large language models always lead to better time-series forecasts than simpler or more task-focused methods. \citep{tan2024llmstimeseries}. Because of that, experiments need to be set up so the benefit from language-model information can be separated from the rest of the changes in the forecasting system.

This work is not set up like methods that plug an LLM or foundation model in as the main prediction engine. Here, the turbine sensor readings are first turned into clear, human-like text that describes the operating state. That text is then mapped into dense vectors that capture meaning. After that, those vectors are added as extra inputs to standard forecasting methods. In this setup, the LLM is used as a semantic context encoder. It is not used to do the forecasting directly. We also run the same forecasting model in two ways. One setup uses no semantic vectors. The other uses the rich semantic features. This No-LLM vs Rich-LLM comparison helps show what the context representation adds on its own.

\subsection{Uncertainty Estimation and Probabilistic Forecasting}

Point forecasts by themselves do not tell you how uncertain future wind power will be. Because of that, wind forecasting studies have turned more often to probabilistic outputs. Examples include prediction intervals, quantiles, and predictive distributions. These tools are used for reserve planning and for operational choices when uncertainty matters \citep{bazionis2022probabilisticwind,xie2023overviewwind}. Conformal prediction presents another possible resolution. This method introduces a methodology to create prediction regions that will have statistically correct coverage. Techniques advanced for time series are also dependent on history, hence in which existence of dependence between the points prevents the standard independent data assumption \citep{xu2021conformal,xu2021enbpi,zaffran2022adaptive}. Such tools are of significant importance in forecast settings to incorporate uncertainties inherent in renewable energy.

Modern time series forecasting models usually have already embraced the probabilistic or distribution oriented approach. A good example of such an approach is the scientific research that respects probability, like Lag-Llama and Chronos models, which are specified for the same purpose. Other examples on the same issue where such methodical approach have been very successful are such models as TimesFM and Moirai which have been trained in large scale forecasts as will be shown by the following articles \citep{rasul2024lagllama,ansari2024chronos,das2024timesfm,woo2024moirai}. Suppose that these modes are relevant to tasks that involve the use of wind power, there are also broader issues at stake. Today, the attention of researchers focuses on making their models forecast in real life conditions and how much uncertainty they carry. And, they are able to be applied to various groups of time series data.

Uncertainty checks may matter even more during ramp events. In those moments, conditions change quickly. They are also less frequent than typical operating states. Because of that, good coverage of prediction intervals for the full test set does not guarantee the same behavior during ramp periods. For this reason, this study looks at prediction interval coverage probability (PICP) and mean prediction interval width (MPIW). We do this for the whole test set. We also report them for ramp-event subsets.

\subsection{Positioning of the Present Study}

Research on wind power forecasting has improved a lot. Progress has been reported on ramp-event prediction, deep sequence modeling, Transformer-based forecasting, time-series foundation models, and probabilistic forecasting. Still, there are open questions that need more work.

First, many wind forecasting papers focus on overall prediction quality. High-ramp cases are less frequent. Yet they matter in day-to-day operations. Because of that, they may need their own evaluation. Second, the papers reviewed for this study, there is not much work on adding LLM-generated semantic notes about turbine operation as extra inputs. Most LLM time-series work either uses the model to predict directly, or treats it as a general time-series encoder. It does not usually use turbine-state descriptions as side features for standard forecasting methods. Third, uncertainty is often checked on the full test set. But how uncertainty behaves during high-ramp operation is not studied as much.

Finally, any statement about the usefulness of LLM-derived signals should be tested against strong non-LLM baselines. This includes classical methods, ensemble systems, and deep sequence learners. Table~\ref{tab:related_work_positioning} places this study in relation to those lines of work. It also lists what this paper adds in each area.

\begin{table}[htbp]
\centering
\caption{Positioning of the present study relative to recent literature.}
\label{tab:related_work_positioning}
\small
\begin{tabularx}{\textwidth}{p{0.22\textwidth} p{0.36\textwidth} X}
\toprule
Research stream & Recent focus & Position of the present study \\
\midrule

Wind power forecasting
& Statistical, machine-learning, deep-learning, hybrid, and probabilistic forecasting
& Evaluates classical, ensemble, sequence-learning, and context-enhanced models across multiple horizons \\

Ramp-event forecasting
& Ramp detection, regression, classification, probabilistic prediction, attention, graph-based models, and multi-task learning
& Defines horizon-specific ramp subsets and evaluates forecasting performance separately during high-ramp conditions \\

Feature representation
& Historical measurements, meteorological inputs, lagged features, statistical features, feature extraction, and learned representations
& Adds symbolic operating states, TF-IDF features, LLM-derived semantic embeddings, and PCA-compressed semantic representations \\

Transformer forecasting
& Attention, decomposition, frequency-domain modeling, patch-based representations, exogenous variables, and multiscale modeling
& Uses strong sequence-learning models as comparison baselines for context-enhanced forecasting \\

LLM and foundation models
& LLM reprogramming, autoregressive LLM forecasting, zero-shot forecasting, and pretrained time-series foundation models
& Uses an LLM as a semantic operating-context encoder rather than as the direct forecasting model \\

Uncertainty estimation
& Probabilistic forecasting, quantile prediction, conformal intervals, and distributional forecasting
& Evaluates interval coverage and width separately for complete-test and ramp-event samples \\

\bottomrule
\end{tabularx}
\end{table}

The idea of this study is focusing on the depicted gaps in the existing literature and employing an advanced semantic approach to delivery of the wind power forecast during ramp events. It consumes wind turbine’s signals, weather measurements and the symbolic operating modes and some auxiliary descriptions in state form that are in short text. Large language model also comes into attention by means of semantic vectors. Tother with ideology testing comes in many forms. Firstly we conduct context ablation checks so see which parts of the strategy work, and which do not. Then we move to comparing models that take account of context and models that do not. Here we consider even more than one forecast model. Some include classical ones, while others are deep sequence models. 

Results come up with bias correction results therefore while using them we present the results by bootstrap confidence intervals. The Diebold-Mariano tests are also implemented. Moreover, the research also examines the sensitivity of the outcomes on the ramp-threshold and looks at this sensitivity across the different definitions of the ramps. We reduce its dimensionality with PCA, which makes the size of the data representation smaller. Furthermore, we perform out-of-sample estimation. Last, we attempt to analyse which combination of factors tends to have significant effects and incorporate an analysis of cautious uncertainties here. In general, the purpose is to identify whether there is any advantage to information incorporation in prognosis due to semantically related context, when this information assists, whether some increase stands as statistically valid, and whether the increase is effective in practice in the case of ramp-event.


\section{Data and Problem Formulation}
\label{sec:data_problem}

\subsection{Dataset Description}

Using the data that was given for the KDD Cup 2022 wind power forecasting challenge, this research employs the methodology of Spatial Dynamic Wind Power Forecasting (SDWPF) with case certified electric power stations. One can see the available data of instantaneous active power provided for 134 wind turbines. Basically, each instant contains a unique turbine identifier, a date and a timestamp and details concerning the current: - velocity and direction of the wind, - temperature, - orientation of the nacelle, - pitch angles of the blades, - resistance or zero inductive power, and - generation of energy. The data to be predicted is the energy output in the form of active power. We denote this quantity as $P_{it}$. Here $i$ indexes the turbine and $t$ is the time period, when we need energy output to serve a predefined task.

For the main analysis, we choose only the turbines from SDWPF whose numbers do not exceed 20. This way we can maintain a multi-turbine forecast configuration. This approach simultaneously will increase the number of repeated model runs, bootstrap sampling, ramp-threshold sensitivity checks, and the LLM-embedding ablation investigations. The measurements were taken every 10 minutes. So the objective is to predict 1, 3 and 6 currencies, i.e. forecasts of 10, 30 and 60 min ahead. For this we also utilize tasks with turbine coordinated feature attributes. These are first normalized before use and they are used to attempt to show what is the volatility in the conditions across the wind farm.

We also conducted experiments on alternative datasets. The first dataset that we refer to is a SCADA data set available on Kaggle which provides information on wind turbine repositories. The second dataset is the one collected by NREL. These additional sets serve as a proof of concept as to whether we can expect improvements with small training set so that there is significant benefit to improving the SCDWPF result. Table~\ref{tab:dataset_setup} provides information on the dataset and the forecasting configuration used.

\begin{table}[htbp]
\centering
\caption{Summary of dataset and forecasting setup.}
\label{tab:dataset_setup}
\small
\begin{tabularx}{\textwidth}{p{0.32\textwidth} X}
\toprule
Item & Description \\
\midrule
Primary dataset 
& Spatial Dynamic Wind Power Forecasting (SDWPF) dataset \\

Original turbine coverage 
& 134 wind turbines \\

Turbines used in main study 
& Turbines 1--20 \\

External validation datasets 
& Kaggle wind turbine SCADA and NREL wind turbine data \\

Sampling interval 
& 10 minutes \\

Forecast horizons 
& 10, 30, and 60 minutes \\

Prediction steps 
& $h \in \{1,3,6\}$ \\

Target variable 
& Future active power output, $P_{i,t+h}$ \\

Main variables 
& Wind speed, wind direction, temperature, nacelle direction, pitch angles, reactive power, and active power \\

Spatial variables 
& Normalized turbine $x$ and $y$ coordinates for SDWPF \\

Data split 
& 70\% training, 15\% validation, and 15\% testing within each turbine sequence \\

Main ramp-event definition 
& Top 10\% of absolute future power changes in the test set for each horizon \\

Sensitivity definitions 
& Top 10\%, 15\%, and 20\% of absolute future power changes \\
\bottomrule
\end{tabularx}
\end{table}

The raw variables are renamed and standardized for consistency. Missing sensor values are forward-filled and backward-filled within each turbine sequence where possible, and missing-sensor indicators are retained. Negative active power values are clipped to zero. Table~\ref{tab:input_variables} summarizes the main input variable groups used in the forecasting framework.

\begin{figure}[htbp]
    \centering
    \includegraphics[width=0.78\textwidth]{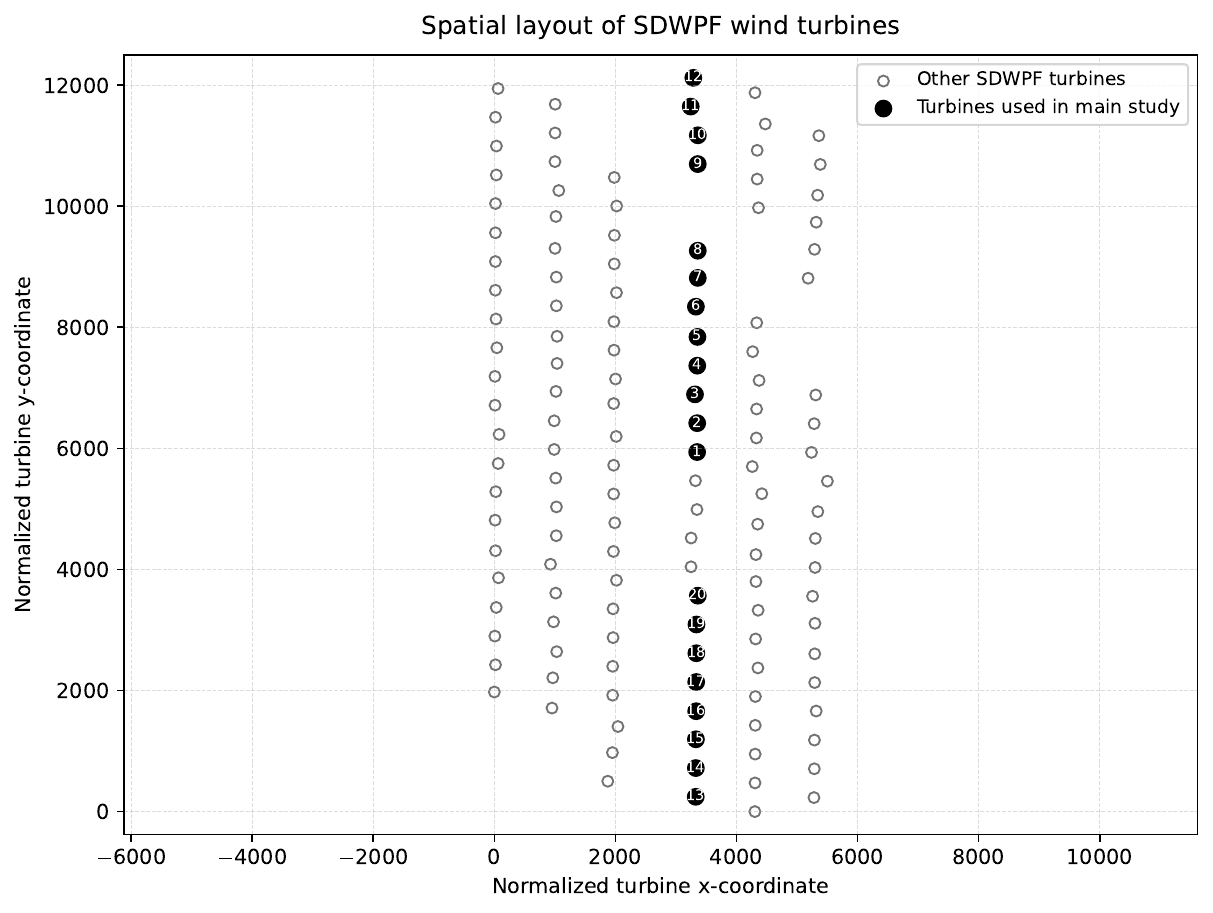}
    \caption{Spatial layout of the SDWPF wind turbines. Filled markers indicate turbines 1--20 used in the main experiments, while open markers indicate the remaining turbines in the wind farm.}
    \label{fig:sdwpf_turbine_layout}
\end{figure}

The SDWPF dataset provides relative turbine coordinates rather than geographic latitude and longitude. Therefore, Figure~\ref{fig:sdwpf_turbine_layout} presents the spatial wind-farm layout instead of a geographic map.

\begin{table}[htbp]
\centering
\caption{Main input variable groups used for wind power forecasting.}
\label{tab:input_variables}
\small
\begin{tabularx}{\textwidth}{p{0.26\textwidth} p{0.36\textwidth} X}
\toprule
Variable group & Variables & Forecasting role \\
\midrule
Meteorological variables 
& Wind speed, wind direction, external temperature 
& Capture atmospheric conditions \\

Turbine operating variables 
& Internal temperature, nacelle direction, blade pitch angles 
& Represent turbine behavior \\

Power variables 
& Active power, reactive power 
& Represent current generation and electrical state \\

Spatial and temporal variables 
& Turbine coordinates, hour, minute, day/month proxies 
& Capture location and time patterns \\

Lagged and rolling variables 
& Power lags, rolling means, rolling standard deviations 
& Capture recent dependence and volatility \\

Context variables 
& Wind-speed change, power change, direction change, variability indicators 
& Capture operating-regime changes \\

Text and semantic variables 
& TF-IDF text features, rich LLM-derived embeddings, PCA-compressed embeddings 
& Encode semantic operating context \\
\bottomrule
\end{tabularx}
\end{table}

\subsection{External Validation Datasets}

Besides the main SDWPF benchmark, we use two outside wind data sources to check whether the semantic operational-context representation can transfer to other settings. One is a Kaggle wind turbine SCADA set. It includes turbine-level signals like active power, wind speed, wind direction, and the theoretical power output. Another collection comes from NREL's wind turbine, which can be downloaded from the NREL's Energy Systems Integration Data Archive. In this instance, the files are in turbine level of stock were used to collect at high frequency in the \texttt{.mat} format. We turn these into 10 points of accumulation of statistics by globalizing active power, wind speed, wind direction, yaw angle, rotor, generator, and blade pitch.

These external datasets do not substitute SDWPF. They act as extra test cases. We want to see if richer features made by the LLM help forecasting more than what the main data already shows. For each external set, we run the same comparison between the paired no-LLM setup and the rich-LLM setup. We do it at 10-, 30-, and 60-minute forecast steps. For NREL, there are fewer ramp-event samples at the long horizons. So we treat the NREL results mostly as a sign that the model stays reliable overall. We do not treat them as the main support for ramp-event claims.

\subsection{Forecasting Problem Formulation}

Let $P_{i,t}$ denote the active power output of turbine $i$ at time step $t$. The objective is to predict future active power at horizon $h$ using information available at time $t$:

\begin{equation}
    \hat{P}_{i,t+h} = f_h(\mathbf{x}_{i,t}),
\end{equation}

where $h \in \{1,3,6\}$ represents the 10-, 30-, and 60-minute horizons, and $\mathbf{x}_{i,t}$ is the input feature vector. Depending on the model specification, $\mathbf{x}_{i,t}$ may include current measurements, lagged values, rolling statistics, spatial and temporal variables, numerical context features, symbolic context features, text features, rich LLM-derived embeddings, and PCA-compressed semantic embeddings.

For each horizon, the supervised learning target is defined as

\begin{equation}
    y_{i,t}^{(h)} = P_{i,t+h}.
\end{equation}

Table~\ref{tab:notation} describes the set of symbols used for designating the times at which a turbine is relevant for observation, the forecast windows that are applied in such observations, the labels where the ramps have occurred, and the subsets for cross-validation. Data is available for each of the turbines in the order of time. The order of the points in the lag as well as the rolling features that are meant to be created is such that they use only the points from the previous period. This stops the leakage from the future that does not exist at the time of the forecast. Each turbine is then split into three groups: the first 70\% is for training, the next 15\% is for validation and the last 15\% is for testing.

\begin{table}[htbp]
\centering
\caption{Mathematical notation used in the forecasting formulation.}
\label{tab:notation}
\begin{tabular}{ll}
\toprule
Notation & Description \\
\midrule
$i$ & Turbine index \\
$t$ & Current time step \\
$h$ & Forecast horizon, where $h \in \{1,3,6\}$ \\
$P_{i,t}$ & Active power output of turbine $i$ at time $t$ \\
$\hat{P}_{i,t+h}$ & Predicted active power output at horizon $h$ \\
$y_{i,t}^{(h)}$ & Forecasting target, defined as $P_{i,t+h}$ \\
$\mathbf{x}_{i,t}$ & Input feature vector available at time $t$ \\
$R_{i,t}^{(h)}$ & Absolute future power change at horizon $h$ \\
$Q_q(R^{(h)})$ & Quantile-based ramp-event threshold for horizon $h$ \\
$\mathbf{z}_{i,t}^{LLM}$ & Rich LLM-derived semantic embedding vector \\
$\mathbf{z}_{i,t}^{PCA}$ & PCA-compressed semantic embedding vector \\
\bottomrule
\end{tabular}
\end{table}

\subsection{Lagged, Rolling, and Context Features}

Historical power information is represented using lagged and rolling features. Active power lags are computed at 1, 2, 3, 6, 12, and 24 time steps:

\begin{equation}
    P_{i,t-k}, \quad k \in \{1,2,3,6,12,24\}.
\end{equation}

Rolling statistics are computed using previous observations only. For a window of length $m$, the rolling mean is

\begin{equation}
    \mu^{P}_{i,t,m}
    =
    \frac{1}{m}
    \sum_{j=1}^{m} P_{i,t-j},
\end{equation}

and the rolling standard deviation is

\begin{equation}
    \sigma^{P}_{i,t,m}
    =
    \sqrt{
    \frac{1}{m-1}
    \sum_{j=1}^{m}
    \left(P_{i,t-j} - \mu^{P}_{i,t,m}\right)^2
    }.
\end{equation}

Rolling means are calculated using the last 3, 6, and 12 time steps. Rolling standard deviations use only the last 6 and 12 steps. To study ramp behavior, we also bring in extra context variables. These include wind-speed change, power change, direction change, power variability, and operating-regime indicators. Those same variables are then used to form symbolic and text-based representations of the operating context.

\subsection{Semantic Operating-Context Representation}

To build the right semantic context, we turn the numeric and symbolic operating states into short natural-language notes. Each note reports how the turbine is doing at time $t$. In particular, it mentions the wind regime, the power regime, how power has changed recently, how wind speed is varying, whether wind direction has shifted, and any ramp-risk context. Then we map those notes into dense semantic embeddings. We call them $\mathbf{z}_{i,t}^{LLM}$. After that, we join these embeddings with the numerical forecasting inputs. This is done in the rich-LLM feature setup:

\begin{equation}
    \mathbf{x}_{i,t}^{RichLLM}
    =
    \left[
    \mathbf{x}_{i,t}^{NoLLM},
    \mathbf{z}_{i,t}^{LLM}
    \right].
\end{equation}

Along with the full 384-dimensional semantic embedding, we also test PCA-compressed versions. For each chosen size $d \in \{16,32,64,128\}$, the PCA semantic embedding is set as

\begin{equation}
    \mathbf{z}_{i,t}^{PCA(d)}
    =
    PCA_d\left(\mathbf{z}_{i,t}^{LLM}\right).
\end{equation}

This check looks at embedding compression. It asks if the meaning can stay in place. At the same time, it tries to cut down noise and reduce the number of dimensions.

\subsection{Ramp-Event Definition}

For a given forecast horizon $h$, the absolute future power change is defined as

\begin{equation}
    R_{i,t}^{(h)} = |P_{i,t+h} - P_{i,t}|.
\end{equation}

The main ramp-event definition uses the top 10\% of $R_{i,t}^{(h)}$ values within the test set for each horizon. A test sample is classified as a ramp event if

\begin{equation}
    R_{i,t}^{(h)} \geq Q_{0.90}\left(R^{(h)}\right),
\end{equation}

where $Q_{0.90}\left(R^{(h)}\right)$ is the 90th percentile threshold for horizon $h$. This percentile-based definition identifies high-change operating periods without requiring a fixed engineering threshold.

\begin{table}[htbp]
\centering
\caption{Main ramp-event definition and horizon-specific thresholds.}
\label{tab:ramp_thresholds}
\small
\begin{tabular}{lccc}
\toprule
Forecast horizon & Prediction step & Ramp definition & Ramp threshold \\
\midrule
10 min & $h=1$ & Top 10\% of $|P_{i,t+1} - P_{i,t}|$ & 93.552 kW \\
30 min & $h=3$ & Top 10\% of $|P_{i,t+3} - P_{i,t}|$ & 172.700 kW \\
60 min & $h=6$ & Top 10\% of $|P_{i,t+6} - P_{i,t}|$ & 248.814 kW \\
\bottomrule
\end{tabular}
\end{table}

To determine whether the outcomes were affected by a single ramp threshold, a further insufficient sensitivity analysis is performed with the initiation of the ramp at the point of the top 10\%, 15\%, and 20\% ramp events. These threshold settings that were utilized within this particular analysis are captured in table VIII. If frequency values are computed at a given quantile level, $q$ within the set $\{0.90,0.85,0.80\}$, then the subset of data is referred to as a gradient boosting.

\begin{equation}
    R_{i,t}^{(h)} \geq Q_q\left(R^{(h)}\right).
\end{equation}

It's important to consider the LNM in the context of these questions and some popular interpretations of the derailment-velocity dependence.

\subsection{Evaluation Subsets}

For each lead time, there are in total two verification datasets. One verification set uses all data available for testing gt sets while the other one, on the other hand, considers the ramp-event samples only. The first set is for the assessment of the overall quality of a particular forecasting configuration. The second veirification set concerns the extent to which the model functions well with rapid changes in the conditions. This distinction is essential. A model might perform sufficiently over the dataset but failmiserably during the ramps which are signal operations. This breaks down for easy determination during the uncertainty splits. In this manner we may know whether prediction limits will hold up in ramp-event hours.


\section{Methodology}
\label{sec:methodology}

\subsection{Overview of the Proposed Framework}

The framework improves typical wind power forecasting by adding numeric, symbolic, text, and meaning based views of the operating context. For turbine $i$ at time $t$, the model pulls in sensor readings. It also takes lagged power data and rolling statistics. In addition, it uses time related and space related variables. On top of that, it feeds in context representations. With all of these inputs, the model predicts the next active power, $P_{i,t+h}$. The forecast horizons are $h \in \{1,3,6\}$. These map to 10, 30, and 60 minutes ahead.

The general forecasting function is written as

\begin{equation}
    \hat{P}_{i,t+h}
    =
    f_h
    \left(
    \mathbf{x}^{num}_{i,t},
    \mathbf{x}^{lag}_{i,t},
    \mathbf{x}^{ctx}_{i,t},
    \mathbf{z}^{txt}_{i,t},
    \mathbf{z}^{LLM}_{i,t}
    \right),
\end{equation}

$\mathbf{x}^{num}_{i,t}$ holds the original numbers from turbine signals, weather, and time and location cues. $\mathbf{x}^{lag}_{i,t}$ holds older power values, plus rolling summaries made from the recent past. $\mathbf{x}^{ctx}_{i,t}$ includes both numeric cues and symbolic context. $\mathbf{z}^{txt}_{i,t}$ is the sparse text-related features. $\mathbf{z}^{LLM}_{i,t}$ is the dense semantic embeddings built from the operational text summaries.

This study does not treat an LLM as a pure forecaster. Instead, it uses the language model to capture meaning from the turbine data. The raw turbine states, which are numbers, are rewritten as short operational descriptions in plain language. Those summaries are then turned into dense vectors. After that, the vectors are merged with the usual forecasting inputs. In this way, the forecast model keeps the useful signal from standard machine learning, but it also benefits from the semantic cues about how each turbine is operating.

\subsection{Numerical and Context Feature Construction}

The numeric feature vector has turbine variables, weather data, time related values, location coordinates, prior active power readings, and summary stats from windows. The lagged power features are defined as

\begin{equation}
    \mathbf{x}^{lag}_{i,t}
    =
    [
    P_{i,t},
    P_{i,t-1},
    P_{i,t-2},
    P_{i,t-3},
    P_{i,t-6},
    P_{i,t-12},
    P_{i,t-24}
    ].
\end{equation}

 means and standard deviations are built from past data only, so no future info can leak into the calculation. These values help show how things keep going over time, what the recent direction looks like, and how much the signal wiggles in the short run.

The ramp numeric context features come from nearby changes in wind speed, active power, and wind direction:

\begin{equation}
    \Delta W_{i,t} = W_{i,t} - W_{i,t-1},
\end{equation}

\begin{equation}
    \Delta P_{i,t} = P_{i,t} - P_{i,t-1},
\end{equation}

\begin{equation}
    \Delta D_{i,t}
    =
    \left[
    \left(D_{i,t} - D_{i,t-1} + 180\right)
    \bmod 360
    \right] - 180.
\end{equation}

The circular transformation limits the direction shift to $[-180^\circ,180^\circ]$. Extra roll wind-speed and power stats are added to show local changes. Together, these details capture the quick operating swings that can suggest ramp-prone periods.

\subsection{Symbolic, Textual, and Semantic Context Representations}

Numeric context variables get split into symbolic operating states. These states cover wind regime, power regime, wind speed change, power output change, wind direction change, and ramp risk. A symbolic context vector is then defined as

\begin{equation}
    \mathbf{c}_{i,t}
    =
    [
    c^{W}_{i,t},
    c^{P}_{i,t},
    c^{\Delta W}_{i,t},
    c^{\Delta P}_{i,t},
    c^{\Delta D}_{i,t},
    c^{R}_{i,t}
    ].
\end{equation}

For sparse text context, symbolic states are concatenated into compact descriptions and transformed into TF-IDF vectors:

\begin{equation}
    \mathbf{z}^{tfidf}_{i,t}
    =
    \phi_{tfidf}
    \left(
    T^{short}_{i,t}
    \right).
\end{equation}

To give better meaning, turbine states shown as numbers and symbols are turned into plain-language summaries of what is happening in operation.

\begin{equation}
    T^{rich}_{i,t}
    =
    g
    \left(
    \mathbf{s}_{i,t},
    \mathbf{x}^{lag}_{i,t},
    \mathbf{x}^{ctx}_{i,t},
    \mathbf{c}_{i,t}
    \right),
\end{equation}

Let $g(\cdot)$ be a fixed text-generation function. The summary it produces covers the wind speed level, the active power level, the latest change in wind speed, the change in power output, a shift in wind direction, short-term variability, what time of day it is, and the ramp-risk state. This detailed context is turned into vectors with a pretrained sentence embedding model:

\begin{equation}
    \mathbf{z}^{LLM}_{i,t}
    =
    \phi_{LLM}
    \left(
    T^{rich}_{i,t}
    \right).
\end{equation}

The resulting rich semantic feature vector is

\begin{equation}
    \mathbf{x}^{RichLLM}_{i,t}
    =
    [
    \mathbf{x}^{NoLLM}_{i,t},
    \mathbf{z}^{LLM}_{i,t}
    ].
\end{equation}

To evaluate whether the full embedding representation introduces unnecessary noise, PCA-compressed embedding variants are also constructed. For $d \in \{16,32,64,128\}$, the compressed semantic representation is

\begin{equation}
    \mathbf{z}^{PCA(d)}_{i,t}
    =
    PCA_d
    \left(
    \mathbf{z}^{LLM}_{i,t}
    \right).
\end{equation}

The corresponding PCA-enhanced feature vector is

\begin{equation}
    \mathbf{x}^{PCA(d)}_{i,t}
    =
    [
    \mathbf{x}^{NoLLM}_{i,t},
    \mathbf{z}^{PCA(d)}_{i,t}
    ].
\end{equation}

Table~\ref{tab:context_representations} lists the context representations we tested. They go from standard numeric features to symbolic ones. We also include text based representations, features from an LLM, and semantic vectors reduced with PCA.
\begin{table}[htbp]
\centering
\caption{Context representations constructed in this study.}
\label{tab:context_representations}
\small
\begin{tabularx}{\textwidth}{p{0.25\textwidth} p{0.38\textwidth} X}
\toprule
Context representation & Construction & Forecasting role \\
\midrule
Numerical context 
& Wind-speed change, power change, direction change, and variability features 
& Captures short-term operating changes \\

Symbolic context 
& Discretized wind, power, change, and ramp-risk categories 
& Represents operating regimes in categorical form \\

TF-IDF text context 
& Short descriptions created from symbolic states 
& Tests sparse textual context representation \\

Rich semantic context 
& Embeddings of detailed operational summaries 
& Captures semantic turbine-state and ramp-risk information \\

PCA-compressed context 
& Lower-dimensional projection of rich semantic embeddings 
& Tests whether compressed semantic features reduce embedding noise \\
\bottomrule
\end{tabularx}
\end{table}

\subsection{Paired No-LLM and Rich-LLM Forecasting Design}

To test how much the rich semantic embeddings help, we run the forecasting model in two versions. One version uses $\mathbf{z}^{LLM}_{i,t}$, and the other leaves it out. The version without the LLM signal is called the no-LLM model. It is defined as follows:

\begin{equation}
    \hat{P}_{i,t+h}^{NoLLM}
    =
    f_h
    \left(
    \mathbf{x}^{NoLLM}_{i,t}
    \right),
\end{equation}

while the rich semantic-context model is defined as

\begin{equation}
    \hat{P}_{i,t+h}^{RichLLM}
    =
    f_h
    \left(
    \mathbf{x}^{RichLLM}_{i,t}
    \right).
\end{equation}

I will use the unchanged learning technique. What is different is the substitution of a higher level of semantic embeddings only. Thus, I expect the shift in outcomes to be a result of more semantic information and not due to any difference in architecture of the model. I apply the same strategy when working with the SDWPF dataset as well. I have also tested this approach on the Kaggle SCADA data, and on the NREL validation dataset.

\subsection{Forecasting Models}

Several forecasting model families are evaluated. The persistence baseline is

\begin{equation}
    \hat{P}_{i,t+h}^{pers} = P_{i,t}.
\end{equation}

The traditional models include moving average in general, moving average with ridge, and moving average with elastic net. These typically utilize historical data of the series and also the power of the series at each window level. When using them against nonlinear techniques, such as machine learning, they can be good reference variables for time series explicit comparative analysis.

In the case of tree and boosting based models, the bagging, random forest, extra trees, histogram-based gradient boosting, extreme gradient boosting, microsoft’s light gradient boosting machine and category boosting are included. Under each of the models, three different settings are defined in terms of feature extraction: the raw-processed feature set, the raw version of the copula feature set, and an SVD-transformed version of the feature set. In default chronicle model, I use two recurrent neural networks (RNNs)-long short-term memory (LSTM) and gated recurrent units (GRUs). Both eat their history and try to predict the action to be made in next step. We aim to investigate how valuable is continuities of the time-series data for prediction of significant events in comparison to the static-typed model collective over such events as well.

For point forecasting models, the standard squared-error objective is

\begin{equation}
    \mathcal{L}_{MSE}
    =
    \frac{1}{N}
    \sum_{n=1}^{N}
    \left(
    y_n - \hat{y}_n
    \right)^2.
\end{equation}

A ramp-weighted variant is also evaluated to increase the influence of high-change samples:

\begin{equation}
    \mathcal{L}_{rw}
    =
    \frac{1}{N}
    \sum_{n=1}^{N}
    w_n
    \left(
    y_n - \hat{y}_n
    \right)^2,
\end{equation}

where

\begin{equation}
    w_n = 1 + \lambda
    \frac{R_n^{(h)}}{\max(R^{(h)})}.
\end{equation}

Here $R_n^{(h)}$ is the absolute future power change and $\lambda$ controls the strength of ramp weighting.

\subsection{Statistical Robustness and Ramp-Threshold Sensitivity}

To check if the gap between the no-LLM and rich-LLM results is real, we run Diebold-Mariano tests. We use squared-error loss for the comparison. For any two forecast methods, we set up the loss gap like this:

\begin{equation}
    d_t
    =
    e_{1,t}^{2}
    -
    e_{2,t}^{2},
\end{equation}

where $e_{1,t}$ and $e_{2,t}$ denote the forecast errors from the two models. If the mean loss difference is above zero, then the second model has smaller squared error on average. The Newey-West adjustment is applied so serial dependence caused by multi-step horizons does not distort the variance.

Bootstrap intervals were formed for the paired RMSE improvements. For every model and each horizon, test set was resampled with replacement. After each resample, recalculation was performed for the RMSE gain that comes from using richer semantic embeddings. Then I take the 2.5th and 97.5th percentiles of those bootstrap results to get the 95\% interval.

In order to perform sensitivity analysis to the ramp thresholds, which will examine the extent of the impact that the selection of different thresholds may have on the outcome, three thresholds were considered, namely the top 10\%, top 15\%, and top 20\% based on the absolute future power change. This is done to see if the findings are simply repeated or if they are robust to different levels of restriction or relaxation on the ramp rule.

\subsection{Uncertainty-Aware Quantile Forecasting}

To assess predictive uncertainty, we use quantile models. They learn three conditional quantiles of future active power. These are for $\tau \in \{0.10,0.50,0.90\}$. The quantile loss is given by

\begin{equation}
    \mathcal{L}_{\tau}
    =
    \frac{1}{N}
    \sum_{n=1}^{N}
    \rho_{\tau}
    \left(
    y_n - \hat{Q}_{\tau,n}
    \right),
\end{equation}

where

\begin{equation}
    \rho_{\tau}(u)
    =
    \begin{cases}
    \tau u, & u \geq 0, \\
    (\tau - 1)u, & u < 0.
    \end{cases}
\end{equation}

The 80\% prediction interval is constructed as

\begin{equation}
    PI_{i,t}^{80\%}
    =
    \left[
    \hat{Q}_{0.10}
    \left(
    P_{i,t+h}
    \mid
    \mathbf{x}_{i,t}
    \right),
    \hat{Q}_{0.90}
    \left(
    P_{i,t+h}
    \mid
    \mathbf{x}_{i,t}
    \right)
    \right].
\end{equation}

Next, estimate the prediction interval all-test set values and ramp-event set values for calculation of the measures Prediction Interval Coverage Decision (PICP) and Prediction Interval Width (MDI) in that order. The aim is to ensure that the calculated uncertainty does not change when the rate of the ramp increases.

\subsection{Feature-Group Importance Analysis}

Tree-based feature importances can be collected into wider classes. For each class, group k holds the related features, written as $\mathcal{G}_k$. Then the share of importance for group k is

\begin{equation}
    S_k
    =
    \frac{
    \sum_{j \in \mathcal{G}_k} I_j
    }{
    \sum_{j=1}^{p} I_j
    }
    \times 100,
\end{equation}

In this setup, $I_j$ means the importance score for feature $j$. Also, $p$ is the total count of features. The feature groups are split into several types: historical power features, physical turbine and weather features, numerical context features, symbolic context features, TF-IDF text features, semantic embedding features, PCA compressed embedding features, and time and location features. With the grouped analysis, we can measure how the effect from semantic operating context embeddings shifts as the forecast horizon changes..

\subsection{Workflow Summary}

Algorithm~\ref{alg:llm_forecasting_framework} summarizes the complete workflow used to generate numerical, symbolic, textual, and semantic representations, train paired no-LLM and rich-LLM models, define ramp-event samples, and evaluate point, statistical, sensitivity, external-validation, and uncertainty results.

\subsection{Workflow Summary}
\label{subsec:workflow_summary}

Algorithm~\ref{alg:llm_forecasting_framework} lays out the full workflow. It covers how we make numerical, symbolic, textual, and semantic forms. It also shows how we train the paired no-LLM model and the rich-LLM model. Next, it describes how ramp-event samples are set up. Finally, it reports the results for point estimates, statistical tests, sensitivity checks, external validation, and uncertainty.

\begin{algorithm}[htbp]
\caption{Hybrid Semantic-Context-Enhanced Ramp Forecasting Framework}
\label{alg:llm_forecasting_framework}

\begin{algorithmic}[1]

\Require Turbine measurements, active power $P_{i,t}$, and horizons
$h \in \{1,3,6\}$

\Ensure Forecasts $\hat{P}_{i,t+h}$, ramp-event metrics, statistical tests, and uncertainty metrics

\State Sort each turbine sequence chronologically
\State Clean missing values and clip negative active power to zero
\State Construct lagged, rolling, temporal, spatial, and numerical context features
\State Convert operating conditions into symbolic context categories
\State Generate natural-language operational summaries from turbine states
\State Encode summaries into rich semantic embeddings
$\mathbf{z}_{i,t}^{\mathrm{LLM}}$
\State Construct PCA-compressed semantic embeddings
$\mathbf{z}_{i,t}^{\mathrm{PCA}(d)}$

\State Form the paired feature sets:
\Statex
\[
\begin{aligned}
\mathbf{x}_{i,t}^{\mathrm{NoLLM}}
&=
\text{numerical and contextual features},
\\
\mathbf{x}_{i,t}^{\mathrm{RichLLM}}
&=
\left[
\mathbf{x}_{i,t}^{\mathrm{NoLLM}},
\mathbf{z}_{i,t}^{\mathrm{LLM}}
\right],
\\
\mathbf{x}_{i,t}^{\mathrm{PCA}(d)}
&=
\left[
\mathbf{x}_{i,t}^{\mathrm{NoLLM}},
\mathbf{z}_{i,t}^{\mathrm{PCA}(d)}
\right].
\end{aligned}
\]

\For{each horizon $h \in \{1,3,6\}$}
    \State Define the forecasting target:
    $y_{i,t}^{(h)} = P_{i,t+h}$

    \State Split each turbine sequence chronologically into training, validation, and testing sets

    \State Train classical, ensemble, boosting, sequence, and quantile models

    \State Define ramp samples using the top 10\%, 15\%, and 20\% of
    $\lvert P_{i,t+h}-P_{i,t} \rvert$

    \State Evaluate full-test and ramp-event forecasting performance

    \State Compute paired LLM gains, Diebold--Mariano tests, and bootstrap confidence intervals
\EndFor

\State Evaluate quantile prediction intervals using PICP and MPIW

\State Repeat the paired evaluation on the external Kaggle SCADA and NREL datasets

\end{algorithmic}
\end{algorithm}

\section{Experimental Setup}
\label{sec:experiments}

\subsection{Forecast Horizons and Evaluation Design}

Tests use three forecast lengths: 10, 30, and 60 minutes. SDWPF is stored every 10 minutes. So those lengths match forecast steps $h \in \{1,3,6\}$. For each $h$, a model is fit separately. The target is the future active power $P_{i,t+h}$. Inputs come only from what is known at time $t$.

There are two evaluation modes. One mode uses every point in the test span. It reports the overall forecast quality. The other mode keeps only ramp-event cases. Here, ramp events are the top 10\% of absolute future power shifts for each horizon. This lets the study compare normal behavior with periods that change fast. Also, ramp-event sensitivity is checked by changing the cutoff. It uses top 10\%, 15\%, and 20\% ramp definitions.

For each turbine sequence, the data are split by time. The first 70\% trains the model. The next 15\% is used to tune it on validation. The last 15\% is held out for testing. This split is chronological, so no future data leaks into training. All lagged and rolling inputs are built only from past values. Table~\ref{tab:evaluation_settings} lists the horizons, how the sets are divided, how ramp events are chosen, which subsets are used for scoring, and which statistical tests are applied.

\begin{table}[htbp]
\centering
\caption{Evaluation settings used in the experiments.}
\label{tab:evaluation_settings}
\small
\begin{tabularx}{\textwidth}{p{0.27\textwidth} p{0.34\textwidth} X}
\toprule
Evaluation setting & Samples included & Purpose \\
\midrule
All-test evaluation 
& All observations in the chronological test period 
& Measures general forecasting performance \\

Ramp-event evaluation 
& Top 10\% highest absolute future power-change samples 
& Measures performance under high-variability regimes \\

Ramp-threshold sensitivity 
& Top 10\%, 15\%, and 20\% highest absolute power-change samples 
& Tests robustness to ramp definition \\

Persistence comparison 
& Model RMSE compared with persistence RMSE 
& Quantifies improvement over a strong short-horizon baseline \\

Paired LLM comparison 
& Same model trained with and without rich semantic embeddings 
& Isolates the contribution of LLM-derived context \\

Bootstrap confidence intervals 
& Resampled paired test predictions 
& Quantifies uncertainty in RMSE gains \\

Diebold-Mariano testing 
& Paired forecast errors from competing models 
& Tests statistical significance of forecast-error differences \\

External validation 
& Kaggle SCADA and NREL wind datasets 
& Tests transferability beyond SDWPF \\

Uncertainty evaluation 
& Quantile prediction intervals on all-test and ramp-event samples 
& Assesses interval reliability under normal and ramp regimes \\
\bottomrule
\end{tabularx}
\end{table}

\subsection{Compared Models}

The comparison between several types of models for forecasting purposes forms the subject of this study. It talks about the traditional time series forecast models, common machine learning (ML) models, models forecasting over sequences of data and XGBoost models extended according to the history of context. Besides this aspect the paper also investigates the impact of the shift from poor LLM models into good LLM models. PCA-compressed contextual meaning is introduced in another subsection. The final part uses quantile models based on Bayesian neural networks that also include the information about uncertainty.

The persistence model has been typically no worse than news forecasting. It uses the active power of the previous period as the reference value to estimate the present and active power consumption in the active power of the following period. In the case of a study of wind power, this model always uses the current and the past power for the two diassociative systems, at adjoining time periods as, as wind power is well-known to show high autocorrelation at short-time intervals, the formulation in the integration group form, or, DYNFEZ is given below. In addition it can complicate the storage system.

There are many types of prediction systems provided by stock market index forecasting. Several different ones have different prediction models, unlike the traditional statistical autoregressive ones. Some primary components of stock index metric time series approaches include: binary forecasts, historical value forecasts, profit and loss, etc. In addition to prediction of direct market index values, two approaches which can potentially constitute aggregated predictions; individual security returns, their variances and convections or measures of systemic markets behaviors are also included in this field of investigation.

Next are the context ablation tests. These rely on XGBoost. The goal is to check how different context forms change results. The variants are XGBoost-Lag, Symbolic Context-XGBoost, TF-IDF TextContext-XGBoost, Short LLMContext-XGBoost, and Rich LLMContext-XGBoost. The check asks whether symbolic, text, or semantic context helps once lagged numeric inputs are already present.

\begin{table}[htbp]
\centering
\caption{Summary of model categories used in the experiments.}
\label{tab:model_configurations}
\small
\begin{tabularx}{\textwidth}{p{0.25\textwidth} p{0.38\textwidth} X}
\toprule
Model category & Models included & Purpose \\
\midrule
Persistence 
& Persistence 
& Strong short-horizon reference baseline \\

Classical time-series baselines 
& Rolling mean, AR-Ridge, AR-ElasticNet 
& Tests autoregressive lag-based forecasting performance \\

Context-ablation XGBoost 
& XGBoost-Lag, Symbolic-XGBoost, TF-IDF-XGBoost, Short LLM-XGBoost, Rich LLM-XGBoost 
& Tests different context representations \\

Expanded ML baselines 
& Ridge, ElasticNet, Random Forest, Extra Trees, HistGradientBoosting, XGBoost, LightGBM, CatBoost 
& Compares against strong non-LLM models \\

Paired LLM comparison 
& Same model trained with and without rich semantic embeddings 
& Isolates the effect of LLM-derived context \\

PCA embedding variants 
& Rich-LLM-384, PCA-16, PCA-32, PCA-64, PCA-128 
& Tests whether compressed semantic embeddings reduce noise \\

Deep sequence baselines 
& LSTM and GRU 
& Provides recurrent sequence-learning comparison \\

Quantile models 
& Quantile XGBoost at $\tau=0.10$, $0.50$, and $0.90$ 
& Produces prediction intervals \\

External validation models 
& Random Forest, Extra Trees, HistGradientBoosting, XGBoost 
& Tests paired LLM gains on Kaggle SCADA and NREL datasets \\
\bottomrule
\end{tabularx}
\end{table}

\subsection{Rich Semantic Embeddings and Paired Comparison}

Semantic-context models use a pretrained sentence-transformer to turn text descriptions into vectors. The short text version keeps things brief and symbolic. The rich text version uses longer operational summaries. These summaries come from the current turbine signals, both numerical and symbolic. After that, the text vectors are added to the feature list. They join the numerical inputs, the lagged values, the rolling stats, and the symbolic features.

In the paired comparison, each selected model family is evaluated under two feature settings:

\begin{equation}
    \mathbf{x}_{i,t}^{NoLLM}
    =
    [
    \mathbf{x}^{num}_{i,t},
    \mathbf{x}^{lag}_{i,t},
    \mathbf{x}^{ctx}_{i,t}
    ],
\end{equation}

\begin{equation}
    \mathbf{x}_{i,t}^{RichLLM}
    =
    [
    \mathbf{x}^{num}_{i,t},
    \mathbf{x}^{lag}_{i,t},
    \mathbf{x}^{ctx}_{i,t},
    \mathbf{z}_{i,t}^{LLM}
    ].
\end{equation}

The paired LLM gain is computed as

\begin{equation}
    Gain_{LLM}(\%)
    =
    \frac{
    RMSE_{NoLLM} - RMSE_{RichLLM}
    }{
    RMSE_{NoLLM}
    }
    \times 100.
\end{equation}

A positive number means that, for the same model type, using richer semantic embeddings lowers the RMSE. On the SDWPF benchmark, most of this improvement shows up in ramp-event RMSE. For the outside Kaggle SCADA data and the NREL data, the paired comparison mostly checks how RMSE holds up across datasets, in general.

\subsection{PCA-Based Embedding Compression}

In order to analyze if all the dimensions in the 384-dimentional structure are useful and none are redundant or noisy, PCA-reduced versions are also tested. These rich semantic embeddings are then preprocessed to 16, 32, 64, and 128 dimensions with the aid of a PCA fitted to the training data. These preprocessed representations are then put together with the no-LLM features and each of the families’ performance is computed.

This PCA comparison helps to study whether it is possible to keep the essence of operational features in a semantically annotated board made out of just a few dimensions. The setups being compared are No-LLM, Rich-LLM-384, Rich-LLM-PCA-16, Rich-LLM-PCA-32, Rich-LLM-PCA-64, and Rich-LLM-PCA-128.

\subsection{Statistical Testing and Bootstrap Confidence Intervals}

To evaluate the significance of any forecast performance improvement that has been noticed, the Diebold-Mariano test is carried out on the paired forecast errors with respect to squared error. The test is performed separately for each forecast horizon and evaluation set. The Newey-West correction is applied to circumvent the dependence of L-step forecast errors resulting from estimated VAR(p) models, and the variance of the forecast accuracy is corrected as well.

Moreover, confidence intervals are produced for the paired no-LLM and rich-LLM RMSE gain proportions without the LL manipulations. For every model and hiss and for every test prediction, the paired test predictions are resampled keeping the number of RMSE gains constant, with the RMSE gain to be computed again withing this resampling. The 2.5th and the 97.5th percentiles of the bootstrapped distribution consistute the 95\% confidence interval. Confidence levels the bar for which reaches zero suggest either significant increases or reductions during resampling depending on the sign.

\subsection{Ramp-Threshold Sensitivity}

The primary ramp-event definition is based on the extreme 10\% increase in future power changes. To eradicate this anchorage method, the study conducted further calibration based on the 15\% and 20\% best power increase in the future, for which, the same calculations for all ramped pairs are executed and recomputed. In conducting the sensitivity examination, the assessment aimed at seeing whether or not the role of textual information is discernible with more or less severe measures of event identification despite defined ramps.

\subsection{External Validation Setup}

In addition to SDWPF benchmark, Kaggle wind turbine SCADA and NREL wind turbine data are taken for the work. These data sets are used to validate the value of intricate semantic operational context beyond the normal framework of SDWPF. Based on the two third-party datasets, the No-LLM vs rich-LLM designs are adopted for 10-, 30-, and 60-minute forecast windows.

Random Forest, Extra Trees, HistGradientBoosting, and XGBoost are among the examined external-validation models. Since the NREL fraction covers very few sample ramp-event samples at the high range of horizons, the external databases serve more as a measure of the overall weather durability rather than the main plank of ramp-event results.

\subsection{Evaluation Metrics}

Point forecasting performance is evaluated using MAE, RMSE, sMAPE, and $R^2$. RMSE improvement over persistence is computed as

\begin{equation}
    \Delta RMSE_{\%}
    =
    \frac{
    RMSE_{pers} - RMSE_{model}
    }{
    RMSE_{pers}
    }
    \times 100.
\end{equation}

When comparing no-LLM and rich-LLM situations, and no-LLM and rich LLMs are combined, RMSE gain is indicated in terms of the percentage reduction in RMSE after the richer semantic embeddings are applied. Moreover, the ramp-threshold sensitivity analysis also estimates the percentage changes in MAE and sMAPE.

Prediction interval coverage probability (PICP) and the mean prediction interval width (MPIW) are important measures if one wishes to conduct evaluations of uncertainty. As a matter of fact, the PICP examines the extent to which the predicted interval contains the observations in terms of percentage, where as the MPIW provides the average width of the intervals. These two are further determined in all-test and ramp-event samples in convenient styles.

\subsection{Implementation Details}

Every analysis was done in Python. The models are very large and applied to XGBoost algorithms. Recurrent neural networks which are well-suited to time-series analysis are mainly implemented with PyTorch with LSTM and GRU as the base models. The origami-like fold model uses encoder models to turn one-tenth of the training data into vectors, and then a decoder model to do surface prediction. Ridge, ElasticNet, Random Forest, Extra Trees, HistGradientBoosting, LightGBM, CatBoost, PCA and bootstrap analyses are performed with available machine learning libraries. All the complex models are trained without consideration for other methods i.e. at a given forecast horizon only.

\section{Results and Discussion}
\label{sec:results}

\subsection{Ramp-Event Forecasting Performance}

This section evaluates the ability of the proposed framework and comparison models to forecast wind power during ramp-event periods. Although persistence is a strong short-horizon baseline for wind power forecasting, its performance deteriorates when active power changes rapidly. Therefore, results are reported separately for ramp-event samples, which correspond to the top 10\% of absolute future power changes at each forecast horizon.

\begin{figure}[htbp]
    \centering
    \includegraphics[width=0.90\textwidth]{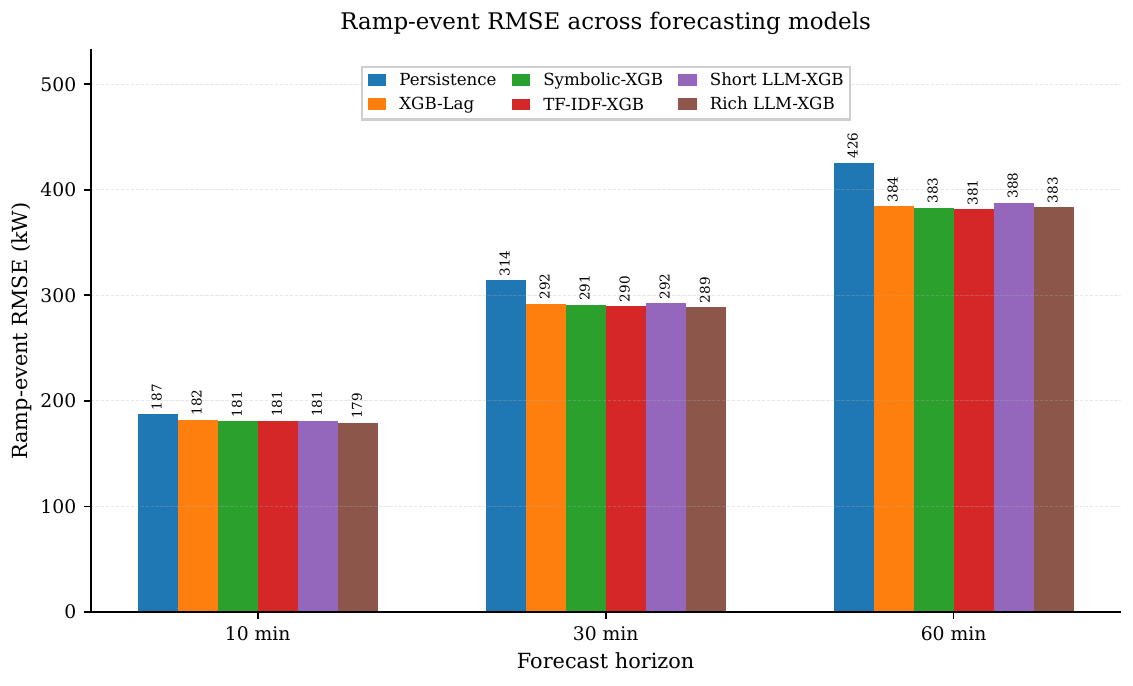}
    \caption{Ramp-event RMSE across context-ablation forecasting models at 10-, 30-, and 60-minute forecast horizons.}
    \label{fig:ramp_rmse_ablation}
\end{figure}

Figure~\ref{fig:ramp_rmse_ablation} compares ramp-event RMSE across the main context-ablation models at the 10-, 30-, and 60-minute forecast horizons.

The context-ablation results show that all context-aware XGBoost models improve over persistence on ramp-event samples. At the 10-minute horizon, persistence obtains a ramp-event RMSE of 187.32 kW, while Rich LLMContext-XGBoost achieves the lowest RMSE among the context-ablation models with 179.03 kW. At the 30-minute horizon, Rich LLMContext-XGBoost again obtains the best context-ablation performance, reducing RMSE from 314.09 kW for persistence to 288.57 kW. At the 60-minute horizon, TF-IDF TextContext-XGBoost achieves the lowest RMSE among the context-ablation models, with 381.36 kW, while Rich LLMContext-XGBoost remains competitive with an RMSE of 383.40 kW.

\begin{figure}[htbp]
    \centering
    \includegraphics[width=0.90\textwidth]{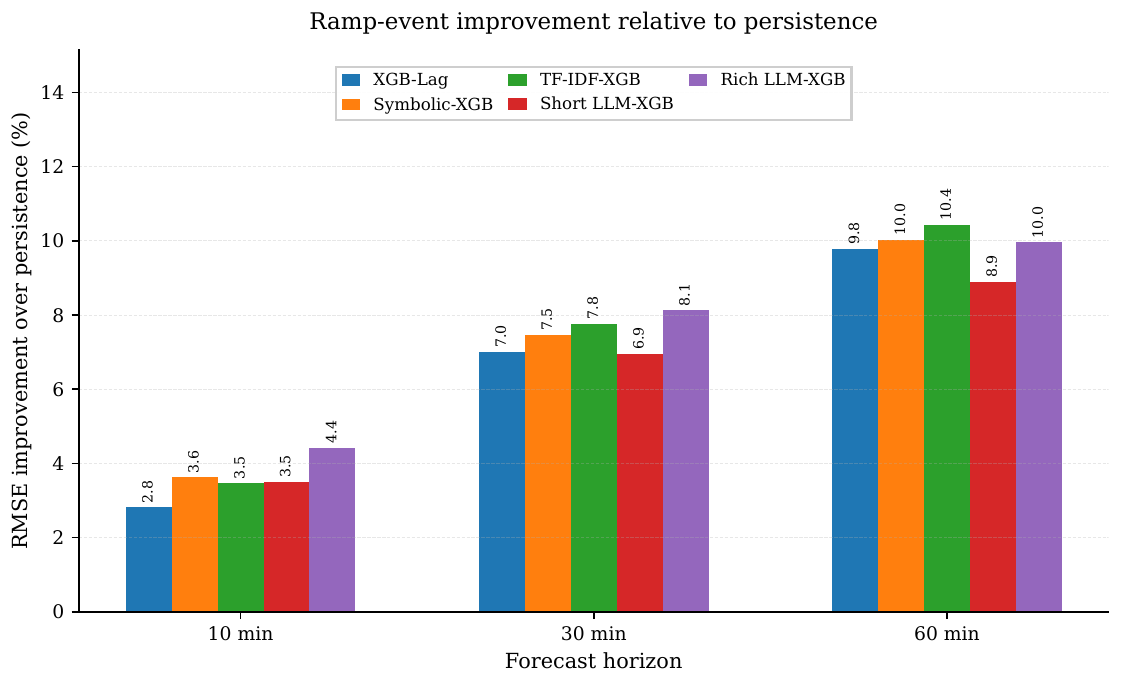}
    \caption{Ramp-event RMSE improvement relative to persistence for context-ablation models. Positive values indicate lower RMSE than persistence.}
    \label{fig:ramp_improvement}
\end{figure}

Figure~\ref{fig:ramp_improvement} reports the corresponding RMSE improvement relative to persistence. Rich LLMContext-XGBoost improves RMSE over persistence by 4.43\%, 8.12\%, and 9.96\% at the 10-, 30-, and 60-minute horizons, respectively. These results show that rich semantic operational context provides value in ramp-event forecasting, particularly at short and medium horizons.

Table~\ref{tab:context_ablation} summarizes the ramp-event ablation results. The ablation evaluates the incremental value of lagged numerical features, symbolic context, TF-IDF text context, short LLM-context embeddings, and rich LLM-context embeddings.

\begin{table}[htbp]
\centering
\caption{Ramp-event forecasting ablation across context representations.}
\label{tab:context_ablation}
\footnotesize
\setlength{\tabcolsep}{4pt}
\renewcommand{\arraystretch}{0.95}
\begin{tabularx}{\textwidth}{p{0.12\textwidth} X rrrr}
\toprule
Horizon & Model & MAE & RMSE & $R^2$ & RMSE gain (\%) \\
\midrule
10 min & Persistence & 167.17 & 187.32 & 0.7439 & 0.00 \\
10 min & XGBoost-Lag & 156.48 & 182.05 & 0.7581 & 2.82 \\
10 min & Symbolic Context-XGBoost & 154.71 & 180.53 & 0.7621 & 3.63 \\
10 min & TF-IDF TextContext-XGBoost & 154.83 & 180.83 & 0.7613 & 3.47 \\
10 min & Short LLMContext-XGBoost & 154.88 & 180.76 & 0.7615 & 3.51 \\
10 min & Rich LLMContext-XGBoost & \textbf{153.77} & \textbf{179.03} & \textbf{0.7660} & \textbf{4.43} \\
\midrule
30 min & Persistence & 286.74 & 314.09 & 0.3104 & 0.00 \\
30 min & XGBoost-Lag & 251.95 & 292.10 & 0.4036 & 7.00 \\
30 min & Symbolic Context-XGBoost & 249.63 & 290.66 & 0.4094 & 7.46 \\
30 min & TF-IDF TextContext-XGBoost & 249.74 & 289.69 & 0.4133 & 7.77 \\
30 min & Short LLMContext-XGBoost & 250.28 & 292.27 & 0.4028 & 6.95 \\
30 min & Rich LLMContext-XGBoost & \textbf{249.83} & \textbf{288.57} & \textbf{0.4179} & \textbf{8.12} \\
\midrule
60 min & Persistence & 396.59 & 425.80 & -0.1874 & 0.00 \\
60 min & XGBoost-Lag & 333.50 & 384.17 & 0.0335 & 9.78 \\
60 min & Symbolic Context-XGBoost & 332.99 & 383.09 & 0.0388 & 10.03 \\
60 min & TF-IDF TextContext-XGBoost & \textbf{330.13} & \textbf{381.36} & \textbf{0.0475} & \textbf{10.44} \\
60 min & Short LLMContext-XGBoost & 336.00 & 387.92 & 0.0144 & 8.90 \\
60 min & Rich LLMContext-XGBoost & 333.52 & 383.40 & 0.0373 & 9.96 \\
\bottomrule
\end{tabularx}
\end{table}

The ablation supports three observations. First, lag-based XGBoost improves over persistence, confirming that nonlinear models using recent power history provide value during high-change periods. Second, symbolic and textual context features provide additional gains, showing that operating-regime information captures structure beyond numerical lags. Third, rich LLM-derived context achieves the best context-ablation performance at the 10- and 30-minute horizons and remains competitive at the 60-minute horizon. Therefore, the ablation does not show that LLM context universally dominates all representations; instead, it shows that rich LLM-derived context is particularly effective at short and medium horizons and remains useful at longer horizons.

\subsection{Statistical Significance of Rich LLM Context}

To assess whether the observed differences between Context-XGBoost and Rich LLMContext-XGBoost are statistically meaningful, a Diebold-Mariano test was conducted using squared-error loss. Positive RMSE gain indicates lower forecast error for the Rich LLMContext-XGBoost model.

\begin{table}[htbp]
\centering
\caption{Diebold-Mariano test results for Context-XGBoost versus Rich LLMContext-XGBoost.}
\label{tab:dm_xgboost_context_llm}
\footnotesize
\setlength{\tabcolsep}{3.5pt}
\renewcommand{\arraystretch}{1.05}
\begin{tabularx}{\textwidth}{
p{0.10\textwidth}
p{0.13\textwidth}
>{\raggedleft\arraybackslash}p{0.14\textwidth}
>{\raggedleft\arraybackslash}p{0.15\textwidth}
>{\raggedleft\arraybackslash}p{0.11\textwidth}
>{\raggedleft\arraybackslash}p{0.11\textwidth}
>{\raggedleft\arraybackslash}X}
\toprule
Horizon & Subset & Context RMSE & Rich LLM RMSE & Gain (\%) & DM stat. & $p$-value \\
\midrule
10 min & All-test   & 67.022  & 65.850  & 1.749  & 15.919 & $<0.001$ \\
10 min & Ramp-event & 180.529 & 179.027 & 0.832  & 6.170  & $<0.001$ \\
30 min & All-test   & 121.869 & 117.211 & 3.822  & 23.477 & $<0.001$ \\
30 min & Ramp-event & 290.656 & 288.570 & 0.718  & 5.067  & $<0.001$ \\
60 min & All-test   & 167.212 & 166.700 & 0.306  & 1.905  & 0.057 \\
60 min & Ramp-event & 383.094 & 383.396 & -0.079 & -0.545 & 0.586 \\
\bottomrule
\end{tabularx}

\vspace{1mm}
\begin{flushleft}
\footnotesize
Note: Positive gain indicates lower RMSE for Rich LLMContext-XGBoost. DM tests use squared-error loss.
\end{flushleft}
\end{table}

As shown in Table~\ref{tab:dm_xgboost_context_llm}, Rich LLMContext-XGBoost significantly improves both all-test and ramp-event forecasts at the 10- and 30-minute horizons. At the 60-minute horizon, the all-test improvement is small and not statistically significant at the 5\% level, while ramp-event performance is slightly worse than Context-XGBoost and also not statistically significant. These findings found that a good deal of contextual information, in general derived from rich lexical-semantic resources, brings about a very substantial statistical improvement over short and medium term time horizons, but the improvement of performance is less evident over the long term and is modulated by the type of representation used.

\subsection{Expanded Model Comparison}

The outcomes of the context ablation analysis reveals the usefulness of the various representations within an XGBoost-based framework. Incomplete. It just does not conform to the expected direction of improvement. Such questions are only answered empirically using large-scale model comparisons. In total 70 models were compared, namely, those based on persistence, regularized linear models, Random Forest, Extra Trees, HistGradientBoosting, XGBoost, LightGBM, and CatBoost. The top 5 ramp-event models at each forecast horizon from the expanded comparison are given in Table~\ref{tab:expanded_top_models}.

\begin{table}[htbp]
\centering
\caption{Top five ramp-event models in the expanded model comparison.}
\label{tab:expanded_top_models}
\footnotesize
\setlength{\tabcolsep}{4pt}
\renewcommand{\arraystretch}{1.05}
\begin{tabularx}{\textwidth}{p{0.10\textwidth} X 
>{\raggedleft\arraybackslash}p{0.11\textwidth}
>{\raggedleft\arraybackslash}p{0.11\textwidth}
>{\raggedleft\arraybackslash}p{0.10\textwidth}
>{\raggedleft\arraybackslash}p{0.13\textwidth}}
\toprule
Horizon & Model & MAE & RMSE & $R^2$ & RMSE gain (\%) \\
\midrule
10 min & XGBoost Standard & 152.731 & 178.848 & 0.767 & 4.524 \\
10 min & Extra Trees & 155.468 & 179.691 & 0.764 & 4.074 \\
10 min & CatBoost & 154.824 & 179.719 & 0.764 & 4.059 \\
10 min & LightGBM & 154.101 & 180.366 & 0.763 & 3.714 \\
10 min & HistGB & 154.435 & 180.433 & 0.762 & 3.678 \\
\midrule
30 min & CatBoost & 249.109 & 287.395 & 0.423 & 8.499 \\
30 min & Random Forest & 247.023 & 288.180 & 0.419 & 8.249 \\
30 min & Extra Trees & 250.502 & 288.858 & 0.417 & 8.033 \\
30 min & ElasticNet & 258.540 & 290.352 & 0.411 & 7.558 \\
30 min & Ridge & 258.568 & 290.363 & 0.411 & 7.554 \\
\midrule
60 min & Random Forest & 323.964 & 376.553 & 0.071 & 11.566 \\
60 min & Ridge & 339.500 & 377.850 & 0.065 & 11.261 \\
60 min & ElasticNet & 339.519 & 377.892 & 0.065 & 11.251 \\
60 min & Extra Trees & 329.017 & 380.046 & 0.054 & 10.745 \\
60 min & CatBoost & 331.174 & 384.307 & 0.033 & 9.745 \\
\bottomrule
\end{tabularx}

\vspace{1mm}
\begin{flushleft}
\footnotesize
Note: HistGB denotes HistGradientBoosting. RMSE gain is measured relative to persistence.
\end{flushleft}
\end{table}

In the broader comparison, it is observed that strong traditional machine learning techniques are effectively effective in the wind power ramp-event prediction problem. At the 10-minute horizon though, XGBoost Standard outperforms these other models, while CatBoost is of relative use at 30-minute horizon, and Random Forest takes the cake at 60 minutes away from the wind ramps considering the non-LLM expanded baselines. This finding is crucial in that it dispels the temptation to exaggerate the benefits of LLMs. For LLMs, it is important to note that this is targeting only particular aspects specific to LLMs and not the more numerically oriented features in the conventional well-calibrated models. In other words, the importance of LLMs should not be lost in that they have to completely eliminate numerical machine learning forecast based models of a high standard in so many possible applications, but rather how much further enhancement it offers in enriching the performance of a model of the same class when scales of language are incorporated as features.

\subsection{Paired No-LLM Versus Rich-LLM Comparison}

To isolate the contribution of rich LLM-derived embeddings, a paired no-LLM versus rich-LLM comparison is conducted. In this experiment, the same model family is trained twice: once without rich LLM embeddings and once with rich LLM embeddings. The difference in ramp-event RMSE is then used to measure the incremental contribution of the LLM-derived representation. Table~\ref{tab:llm_gain_by_model} shows the paired LLM and Non-LLM comparisons for each modeling horizon, which depict the change in RMSE before and after the addition of SESA along with the LLM-based semantic embeddings.

\begin{table}[htbp]
\centering
\caption{Effect of adding rich LLM-derived embeddings to different model families on SDWPF ramp-event RMSE.}
\label{tab:llm_gain_by_model}
\footnotesize
\setlength{\tabcolsep}{4pt}
\renewcommand{\arraystretch}{1.05}
\begin{tabularx}{\textwidth}{p{0.14\textwidth} X 
>{\raggedleft\arraybackslash}p{0.15\textwidth}
>{\raggedleft\arraybackslash}p{0.16\textwidth}
>{\raggedleft\arraybackslash}p{0.14\textwidth}}
\toprule
Horizon & Model & No-LLM RMSE & Rich-LLM RMSE & LLM gain (\%) \\
\midrule
10 min & Random Forest & 179.775 & 178.753 & 0.569 \\
10 min & CatBoost & 179.564 & 178.840 & 0.403 \\
10 min & XGBoost & 179.233 & 178.971 & 0.146 \\
10 min & Extra Trees & 179.339 & 179.463 & -0.069 \\
10 min & HistGB & 180.411 & 180.550 & -0.077 \\
\midrule
30 min & HistGB & 293.396 & 289.110 & 1.461 \\
30 min & XGBoost & 292.608 & 289.286 & 1.135 \\
30 min & Random Forest & 287.640 & 285.328 & 0.804 \\
30 min & CatBoost & 288.334 & 286.473 & 0.645 \\
30 min & Extra Trees & 288.232 & 288.127 & 0.037 \\
\midrule
60 min & CatBoost & 382.290 & 379.094 & 0.836 \\
60 min & XGBoost & 390.024 & 387.990 & 0.522 \\
60 min & HistGB & 390.564 & 389.185 & 0.353 \\
60 min & Extra Trees & 381.328 & 380.150 & 0.309 \\
60 min & Random Forest & 374.710 & 373.682 & 0.274 \\
\bottomrule
\end{tabularx}

\vspace{1mm}
\begin{flushleft}
\footnotesize
Note: HistGB denotes HistGradientBoosting. Positive LLM gain indicates lower ramp-event RMSE after adding rich LLM-derived embeddings.
\end{flushleft}
\end{table}

Through the paired comparison it can be seen that the richer LLM embeddings present improved ramp-event RMSE for most model families. At the 10-minute prediction horizon, three of five models gained upon addition of rich LLM embeddings. This improvement expanded to all evaluated model families’ models at the 30-minute horizon, with the most gain recorded for the HistGradientBoosting model. At the 60-minute horizon, all model families models subjected to evaluation recorded the same improvement, yet the most significant gain was registered by the CatBoost model. Below results exhibit LLM based representation further enhances the solving ability marginally especially for the relatively and the longer length of time horizons.

Although the percentage improvement observed is small, it is statistically significant, as the comparison is controlled and in paired samples. Because the same learning method is applied for the experiments with and without rich LLM embeddings, the difference can be ascribed to morphologization rather than a change in the choice of the model. This provides the argument advanced by the researchers which asserts how in the present market in question, no rich LLM-generated operational context outlasts and fills the gap which is typically occupied by traditional presumption-based models, albeit strengthens them with an ability to provide semantics on a regime basis.

\subsection{Bootstrap Confidence Intervals for LLM Gain}

In order to demonstrate the consistency of the gains of the paired LLMs in the context of sample instability, in addition to the traditional approach of computing confidence intervals, authors also offer an innovative empirical methodology of employing bootstrap methodology to conduct further analysis to support the reader's confidence in the derived results. In this sense, for the ramp-event RMSE gain of each model and horizon, bootstrap confidence intervals were provided and illustrated in the following figure (Figure~\ref{fig:bootstrap_llm_gain}). This resulting representation of the 95\% bootstrap confidence intervals can be observed in Figure~\ref{fig:bootstrap_llm_gain}.

\begin{figure}[htbp]
    \centering
    \includegraphics[width=1.0\textwidth]{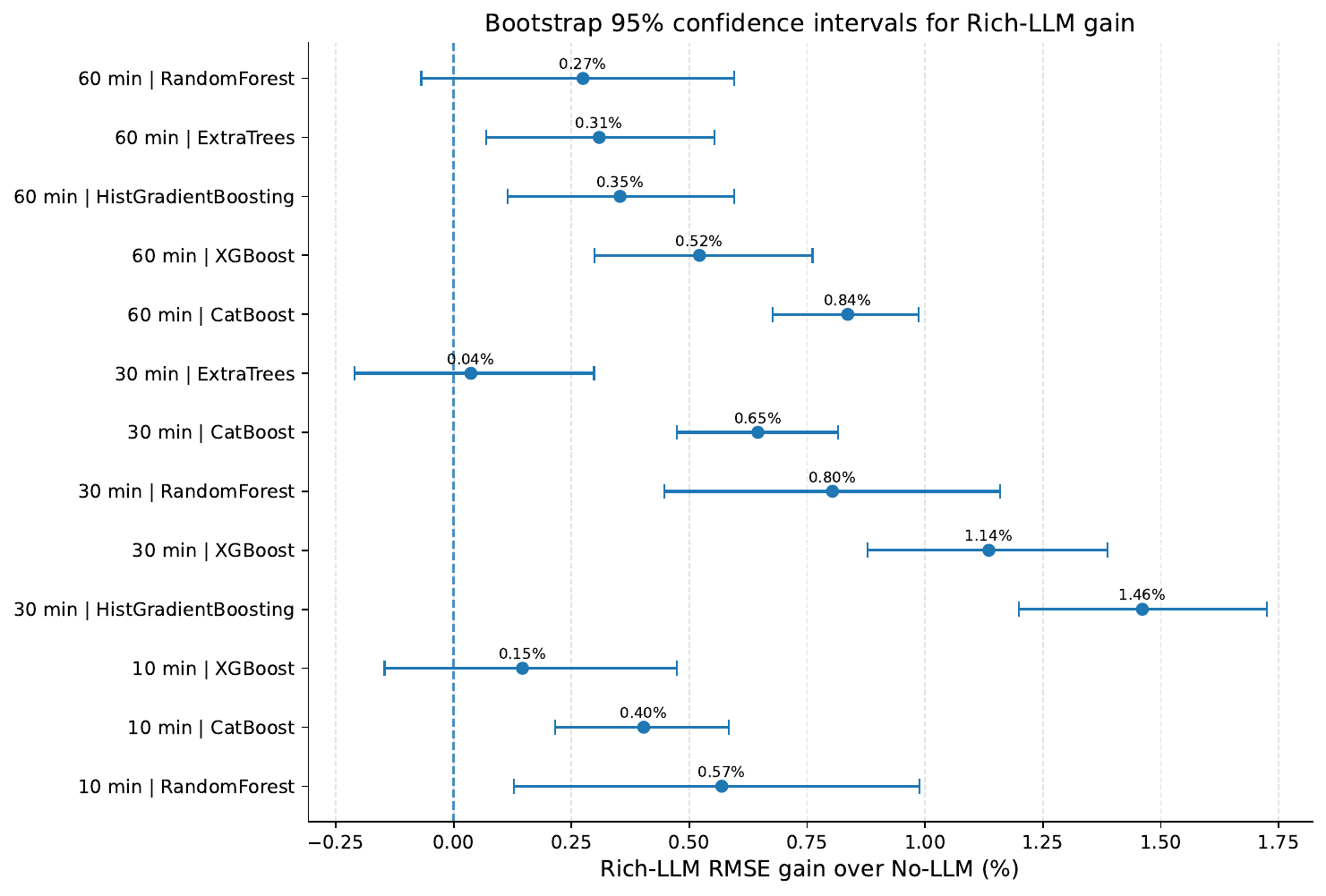}
    \caption{Bootstrap 95\% confidence intervals for Rich-LLM RMSE gain over the corresponding no-LLM model. Positive values indicate lower ramp-event RMSE after adding rich LLM-derived embeddings.}
    \label{fig:bootstrap_llm_gain}
\end{figure}

\begin{table}[htbp]
\centering
\caption{Bootstrap confidence intervals for paired Rich-LLM ramp-event RMSE gain.}
\label{tab:bootstrap_llm_gain}
\footnotesize
\setlength{\tabcolsep}{4pt}
\renewcommand{\arraystretch}{1.05}
\begin{tabularx}{\textwidth}{p{0.13\textwidth} X 
>{\raggedleft\arraybackslash}p{0.15\textwidth}
>{\raggedleft\arraybackslash}p{0.15\textwidth}
>{\raggedleft\arraybackslash}p{0.15\textwidth}}
\toprule
Horizon & Model & RMSE gain (\%) & 95\% CI lower & 95\% CI upper \\
\midrule
10 min & Random Forest & 0.569 & 0.128 & 0.989 \\
10 min & CatBoost & 0.403 & 0.216 & 0.583 \\
10 min & XGBoost & 0.146 & -0.147 & 0.473 \\
\midrule
30 min & HistGB & 1.461 & 1.199 & 1.725 \\
30 min & XGBoost & 1.135 & 0.878 & 1.388 \\
30 min & Random Forest & 0.804 & 0.447 & 1.160 \\
30 min & CatBoost & 0.645 & 0.474 & 0.815 \\
30 min & Extra Trees & 0.037 & -0.211 & 0.298 \\
\midrule
60 min & CatBoost & 0.836 & 0.677 & 0.987 \\
60 min & XGBoost & 0.522 & 0.299 & 0.761 \\
60 min & HistGB & 0.353 & 0.115 & 0.596 \\
60 min & Extra Trees & 0.309 & 0.069 & 0.553 \\
60 min & Random Forest & 0.274 & -0.069 & 0.595 \\
\bottomrule
\end{tabularx}

\vspace{1mm}
\begin{flushleft}
\footnotesize
Note: HistGB denotes HistGradientBoosting. Confidence intervals are based on bootstrap resampling of paired ramp-event predictions.
\end{flushleft}
\end{table}

In Table~\ref{tab:bootstrap_llm_gain}, one can find the bootstrap confidence intervals for the Rich-LLM RMSE gain and whether the improvements hold in the resampled data. It can be observed from the bootstrap results that there are several spending rates, where LLM remains in the positive range in all cases. At the 10-minute horizon – parameters determined by Random Forest and CatBoost possess 95\% confidence intervals which are greater than 0, whereas the terminable node constructed by XGBoost extended beyond 0. Within 30 minutes out of the five model families, four of them show entirely positive confidence intervals of their changes. At the 60-minute horizon, CatBoost, XGBoost, HistGradientBoosting, and Extra Trees also show stable positive gains. These findings strengthen the evidence that the observed LLM improvements are not solely artifacts of a single test sample, although the magnitude of improvement remains modest.

\subsection{Ramp-Threshold Sensitivity}

\begin{figure}[htbp]
    \centering
    \includegraphics[width=0.90\textwidth]{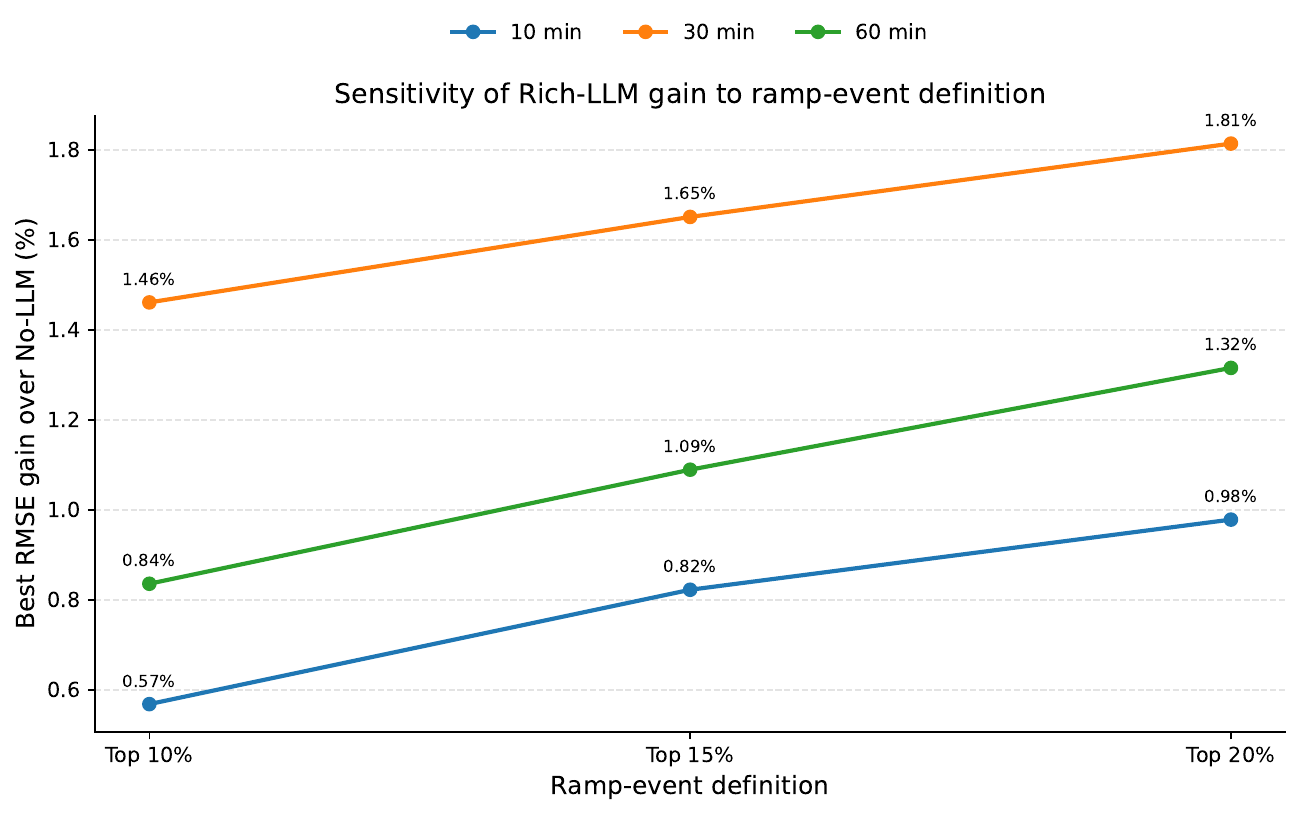}
    \caption{Sensitivity of best Rich-LLM RMSE gain to ramp-event definition. Positive values indicate lower RMSE after adding rich LLM-derived embeddings.}
    \label{fig:ramp_threshold_sensitivity}
\end{figure}

The main ramp-event definition uses the top 10\% of absolute future power changes. To test whether the conclusions depend on this single threshold, additional ramp definitions are evaluated using the top 15\% and top 20\% of absolute future power changes. As shown in Figure~\ref{fig:ramp_threshold_sensitivity}, the Rich-LLM representation continues to provide positive RMSE gains under alternative ramp definitions, indicating that the observed improvements are not limited to a single threshold choice.

Table~\ref{tab:ramp_threshold_sensitivity} summarizes the Rich-LLM RMSE gains under the top 10\%, 15\%, and 20\% ramp-event definitions, showing whether the conclusions are sensitive to the selected ramp threshold.
\begin{table}[htbp]
\centering
\caption{Ramp-threshold sensitivity of the best Rich-LLM RMSE gain.}
\label{tab:ramp_threshold_sensitivity}
\small
\setlength{\tabcolsep}{5pt}
\renewcommand{\arraystretch}{1.05}
\begin{tabular}{lllr}
\toprule
Horizon & Ramp definition & Best model & RMSE gain (\%) \\
\midrule
10 min & Top 10\% & Random Forest & 0.569 \\
10 min & Top 15\% & Random Forest & 0.823 \\
10 min & Top 20\% & Random Forest & 0.979 \\
\midrule
30 min & Top 10\% & HistGB & 1.461 \\
30 min & Top 15\% & HistGB & 1.651 \\
30 min & Top 20\% & XGBoost & 1.814 \\
\midrule
60 min & Top 10\% & CatBoost & 0.836 \\
60 min & Top 15\% & CatBoost & 1.089 \\
60 min & Top 20\% & CatBoost & 1.316 \\
\bottomrule
\end{tabular}

\vspace{1mm}
\begin{flushleft}
\footnotesize
Note: HistGB denotes HistGradientBoosting.
\end{flushleft}
\end{table}

The sensitivity analysis shows that the positive effect of rich LLM-derived context is not limited to a single ramp-event definition. Across all horizons, the best LLM gain remains positive under top 10\%, 15\%, and 20\% ramp definitions. The gains generally become larger when the ramp definition is relaxed from the most extreme top 10\% samples to the broader top 20\% samples. This suggests that semantic context is especially useful for high-change operating regimes, but its benefit may be more stable when the ramp sample includes a broader range of high-variability conditions.

\subsection{PCA Compression of Rich LLM Embeddings}

The rich LLM representation uses a 384-dimensional sentence embedding. To determine whether this full representation is always necessary, PCA-compressed variants are evaluated at 16, 32, 64, and 128 dimensions. Table~\ref{tab:pca_summary} reports the best semantic representation for each model and horizon.

\begin{table}[htbp]
\centering
\caption{Summary of Rich-LLM and PCA-compressed semantic embedding performance.}
\label{tab:pca_summary}
\footnotesize
\setlength{\tabcolsep}{4pt}
\renewcommand{\arraystretch}{1.08}
\begin{tabularx}{\textwidth}{
p{0.10\textwidth}
p{0.18\textwidth}
p{0.22\textwidth}
>{\raggedleft\arraybackslash}p{0.13\textwidth}
>{\raggedleft\arraybackslash}X}
\toprule
Horizon & Model & Best semantic variant & Best RMSE & Gain vs. No-LLM (\%) \\
\midrule
10 min & Random Forest & Rich-LLM-384 & 179.795 & 1.334 \\
10 min & HistGB & Rich-LLM-384 & 180.316 & 0.553 \\
10 min & XGBoost & PCA-64 & 180.323 & 0.386 \\
10 min & CatBoost & Rich-LLM-384 & 179.159 & 0.295 \\
\midrule
30 min & Random Forest & Rich-LLM-384 & 285.783 & 1.568 \\
30 min & XGBoost & Rich-LLM-384 & 288.725 & 1.077 \\
30 min & HistGB & Rich-LLM-384 & 288.553 & 0.731 \\
30 min & CatBoost & PCA-32 & 285.509 & 0.365 \\
30 min & Extra Trees & PCA-128 & 288.102 & 0.228 \\
\midrule
60 min & XGBoost & PCA-64 & 389.485 & 2.086 \\
60 min & Random Forest & PCA-64 & 373.891 & 1.023 \\
60 min & HistGB & PCA-128 & 385.805 & 0.972 \\
60 min & CatBoost & Rich-LLM-384 & 378.361 & 0.684 \\
60 min & Extra Trees & PCA-128 & 378.780 & 0.512 \\
\bottomrule
\end{tabularx}

\vspace{1mm}
\begin{flushleft}
\footnotesize
Note: HistGB denotes HistGradientBoosting.
\end{flushleft}
\end{table}

The PCA analysis shows that the full 384-dimensional rich LLM representation is often beneficial, especially at the 10- and 30-minute horizons. However, compressed embeddings sometimes perform better, particularly at the 60-minute horizon. The strongest PCA result occurs for XGBoost at the 60-minute horizon, where PCA-64 reduces RMSE by 2.086\% relative to the no-LLM model. This means that in some cases with a higher time horizon, reducing dimensionality may help to reduce the amount of noise in a semantic representation. In this case, the embedding-ablation findings are in favor of the use of semantic context and at the same time, emphasize the fact that the increase in the number of embedding dimensions does not always help in that.

\subsection{Final Model Comparison with Classical and Deep Sequence Baselines}

The elaborated analyses cover the areas of the context ablation, the gains in case of two combined LLMs, the bootstrap stability, the ramp-threshold sensitivity and the embedding compression. These results range further complex forecasting tasks and are put into perspective in comparison to Table~\ref{tab:final_model_comparison} where the best rich-LLM ensemble model is contrasted with the classical autoregressive baselines and deep sequence models.

\begin{table}[htbp]
\centering
\caption{Final ramp-event model comparison including classical, ensemble, and deep sequence baselines.}
\label{tab:final_model_comparison}
\footnotesize
\setlength{\tabcolsep}{4pt}
\renewcommand{\arraystretch}{1.05}
\begin{tabularx}{\textwidth}{p{0.10\textwidth} X
>{\raggedleft\arraybackslash}p{0.11\textwidth}
>{\raggedleft\arraybackslash}p{0.11\textwidth}
>{\raggedleft\arraybackslash}p{0.11\textwidth}
>{\raggedleft\arraybackslash}p{0.10\textwidth}}
\toprule
Horizon & Model group & MAE & RMSE & sMAPE & $R^2$ \\
\midrule
10 min & Rich-LLM ensemble & 154.555 & \textbf{178.753} & 30.730 & 0.767 \\
10 min & No-LLM ensemble & \textbf{152.766} & 179.233 & 31.060 & 0.766 \\
10 min & GRU & 158.040 & 182.352 & 32.448 & 0.757 \\
10 min & LSTM & 154.636 & 182.350 & \textbf{30.854} & 0.757 \\
10 min & Persistence & 167.170 & 187.323 & -- & 0.744 \\
10 min & AR-Ridge & -- & 202.601 & -- & -- \\
10 min & AR-ElasticNet & -- & 202.497 & -- & -- \\
\midrule
30 min & GRU & \textbf{243.427} & \textbf{284.557} & \textbf{46.989} & \textbf{0.433} \\
30 min & Rich-LLM ensemble & -- & 285.328 & -- & -- \\
30 min & No-LLM ensemble & -- & 287.640 & -- & -- \\
30 min & LSTM & 248.581 & 291.653 & 48.356 & 0.404 \\
30 min & AR-ElasticNet & -- & 296.926 & -- & -- \\
30 min & AR-Ridge & -- & 296.936 & -- & -- \\
30 min & Persistence & 286.740 & 314.090 & -- & 0.310 \\
\midrule
60 min & GRU & \textbf{320.140} & \textbf{368.729} & \textbf{59.274} & \textbf{0.105} \\
60 min & Rich-LLM ensemble & -- & 373.682 & -- & -- \\
60 min & No-LLM ensemble & -- & 374.710 & -- & -- \\
60 min & LSTM & 325.006 & 375.152 & 61.117 & 0.073 \\
60 min & AR-Ridge & -- & 389.285 & -- & -- \\
60 min & AR-ElasticNet & -- & 389.319 & -- & -- \\
60 min & Persistence & 396.590 & 425.799 & -- & -0.187 \\
\bottomrule
\end{tabularx}

\vspace{1mm}
\begin{flushleft}
\footnotesize
Note: Rich-LLM ensemble denotes the best ensemble model using rich LLM-derived embeddings. No-LLM ensemble denotes the best corresponding ensemble model without LLM-derived embeddings. Dashes indicate metrics that were not available from the corresponding baseline experiment.
\end{flushleft}
\end{table}

The final comparison is conclusive about one thing - that rich-LLM ensemble models are effective; in other more critical terms, they do not perform the best in all model categories. According to the 10-minute case, the rich-LLM ensemble approach emerges as the one which gives the lowest RMSE on slope failure. On the other hand, with respect to 30 and 60 and minute cases, GRU provides the best approach in terms of RMSE on the slope failure most probably since it is capable of retaining memory in the recurrent blobs for longer horizon prediction (ramp). Yet, the poor-LLM ensemble models are also useful and perform better in comparison with the most suitable no-LLM ensemble models. This effect between all the preceding comparative cases supports the generalized mall part of the report: the derived through lllm methodology part of the semantic context is put forward as a supplementary customer behaviour prediction but not in the manner of completely replacing it. For an effervescent discussion, the scores of the lllm semantic context are contrasted with that of some commonly used benchmarks and with a few esoteric and obscure s-e baselines as well.

\subsection{External Validation Results}

To examine the feasibility of implementing the advanced models for operational forecasting in an environment different from the SDWPF benchmark, particularly generic rich-LLM approach, two additional external wind time series are employed: the SCADA data provided by Kaggle and the NREL wind turbine data. For each of the datasets, the same no LLM versus rich LLM design is enforced for 10, 30, and 60-minute forecast horizons. The external precision and bias tests are mostly oriented at the overall RMSE because the NREL dataset has low numbers of ramp event observations for the longer horizon. Table~\ref{tab:external_validation} presents the external validation scores with respect to the other datasets which illustrates the dearth of the event-type data that can be of longer horizon to the RMSE measure for a complete given problem, in the case of the NREL subset. Comparing time horizons, different NSLR values and best non-LLM’ individual results have been described in the enhanced NDF civil judgment prediction models.

\begin{table}[htbp]
\centering
\caption{External validation of best non-LLM and best LLM-augmented models. Positive gain indicates lower overall RMSE for the LLM-augmented model.}
\label{tab:external_validation}
\footnotesize
\setlength{\tabcolsep}{4pt}
\renewcommand{\arraystretch}{1.05}
\begin{tabularx}{\textwidth}{
p{0.17\textwidth}
p{0.11\textwidth}
>{\raggedleft\arraybackslash}p{0.15\textwidth}
>{\raggedleft\arraybackslash}p{0.14\textwidth}
X
>{\raggedleft\arraybackslash}p{0.12\textwidth}}
\toprule
Dataset & Horizon & Non-LLM RMSE & LLM RMSE & Best LLM model & LLM gain (\%) \\
\midrule
Kaggle SCADA & 10 min & 262.889 & 262.502 & HistGB & 0.147 \\
Kaggle SCADA & 30 min & 419.826 & 417.090 & Random Forest & 0.652 \\
Kaggle SCADA & 60 min & 564.367 & 554.269 & Extra Trees & 1.789 \\
NREL Wind & 10 min & 10.354 & 10.046 & Extra Trees & 2.967 \\
NREL Wind & 30 min & 15.084 & 13.483 & Extra Trees & 10.615 \\
NREL Wind & 60 min & 20.755 & 18.305 & Extra Trees & 11.805 \\
\bottomrule
\end{tabularx}

\vspace{1mm}
\begin{flushleft}
\footnotesize
Note: HistGB denotes HistGradientBoosting. The table reports overall RMSE on external validation datasets.
\end{flushleft}
\end{table}

\begin{figure}[htbp]
    \centering
    \includegraphics[width=0.90\textwidth]{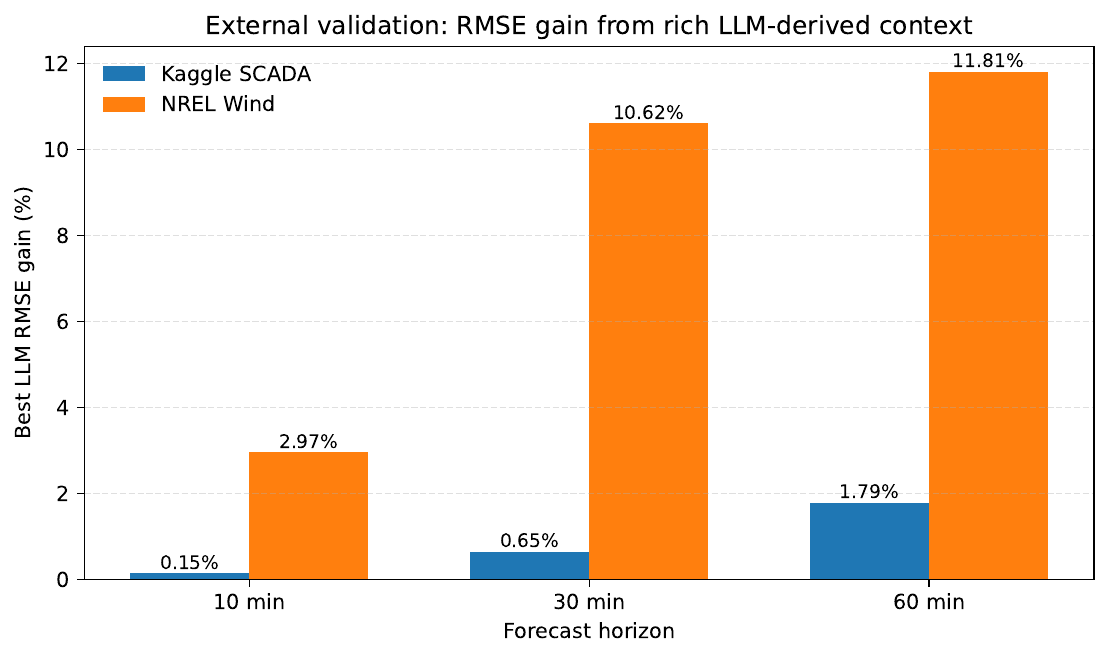}
    \caption{External validation results showing overall RMSE gain from adding rich LLM-derived operational context on Kaggle SCADA and NREL wind datasets. Positive values indicate lower RMSE for the LLM-augmented model compared with the best non-LLM counterpart.}
    \label{fig:external_validation_llm_gain}
\end{figure}

Table~\ref{tab:external_validation} and Figure~\ref{fig:external_validation_llm_gain} clearly demonstrate that the best-performing LLM-augmented models decrease the overall RMSE for all the evaluation intervals on the external dataset. The results provide modest but steady improvements in the case of Kaggle SCADA, from 0.147\% at the 10-minute horizon to 1.789\% at the 60-minute horizon. For the NREL wind dataset, the results are even better, starting at 2.967\% at the 10-minute horizon and rising to 11.805\% at the 60-minute horizon. These results suggest that the developed rich LLM-based operational context may be applicable for a longer forecasting period as well.

The external validation also shows that the magnitude of LLM benefit varies across datasets. The Kaggle SCADA gains are smaller, suggesting that the numerical turbine variables already capture much of the predictive structure. In contrast, the larger NREL gains indicate that semantic operating-context embeddings can provide stronger incremental value when the baseline feature representation is less expressive or when operating regimes differ more strongly across the forecasting horizon. However, external ramp-event results should be interpreted cautiously. Kaggle SCADA contains sufficient ramp-event samples, whereas the NREL subset contains only 9, 2, and 1 ramp-event samples at the 10-, 30-, and 60-minute horizons, respectively. Therefore, the external datasets are used mainly to support general forecasting robustness, while the SDWPF benchmark remains the primary basis for ramp-event evaluation.

\subsection{Rich LLM Feature Contribution}

\begin{figure}[htbp]
    \centering
    \includegraphics[width=0.90\textwidth]{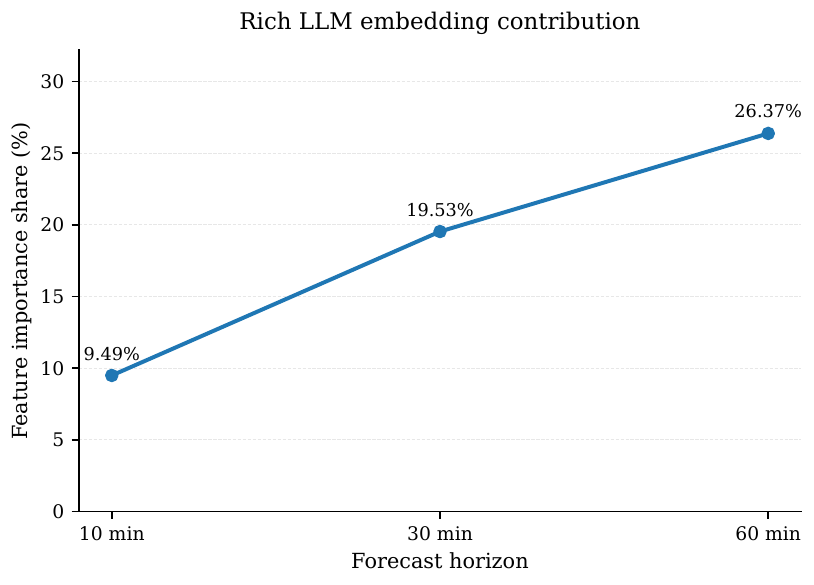}
    \caption{Feature-importance contribution of rich LLM-context embeddings across forecast horizons.}
    \label{fig:llm_importance}
\end{figure}

To determine whether rich LLM-context embeddings are actually used by the forecasting model, feature-group importance analysis is performed. Feature importance values from the XGBoost models are grouped into broader categories, including historical power features, physical turbine and weather features, numerical context features, symbolic context features, text features, and rich LLM embedding features. Figure~\ref{fig:llm_importance} shows the feature-importance share of rich LLM embeddings across forecast horizons. The contribution of rich LLM embeddings increases from 9.49\% at the 10-minute horizon to 19.53\% at the 30-minute horizon and 26.37\% at the 60-minute horizon.

The significance of the contribution by recent active power and numerical lags at very short horizons is that they still remain highly informative even to strong rheological time series autocorrelation. For the most part, however, the system focuses on the semantic context that captures and combines the information about the situation, describing it in a general way, such as wind speed changes or unstable power generation due to shifting wind erectors, and any debates about the perceived state of risk concerning ramping. This suggests that a representation derived from LLM is not just another feature for the forecast model but is helpful in the process.

\subsection{Uncertainty-Aware Forecasting Results}

Aside from equivalent level forecasting, an assessment of uncertainty is done by utilizing a Rich LLMContext quantile XGBoost model. The model calculates the 0.10, 0.50, and 0.90 conditional quantiles of active power coming, which leads to the construction of an 80\% prediction interval. The prediction interval coverage probability and the mean forecast interval width are examined providing separate analysis for all-test and ramp-event samples.

\begin{figure}[htbp]
    \centering
    \includegraphics[width=0.80\textwidth]{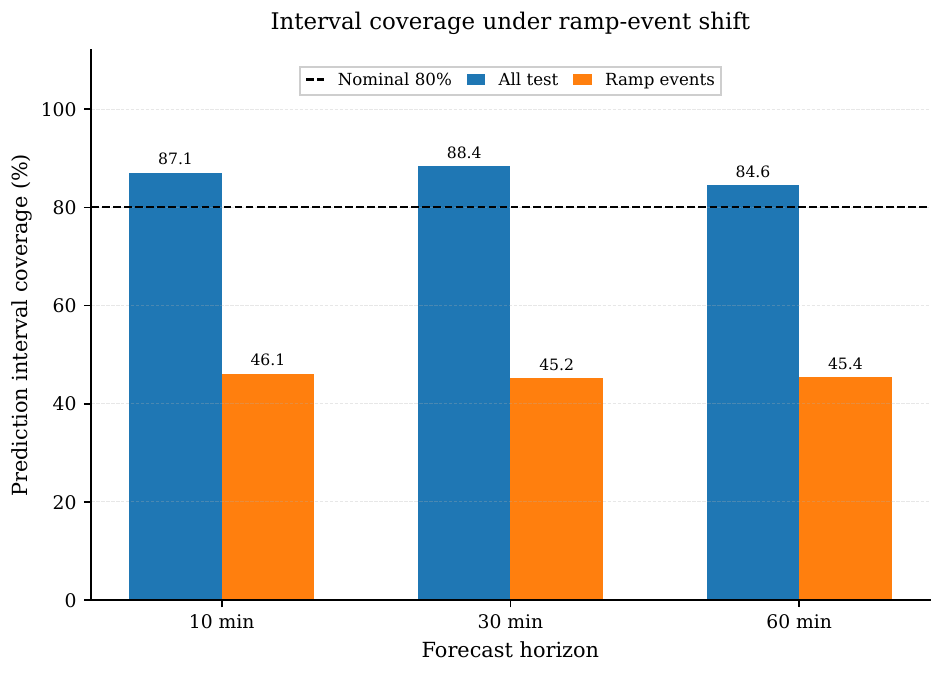}
    \caption{Prediction interval coverage probability for all-test and ramp-event samples. The nominal prediction interval is 80\%.}
    \label{fig:picp}
\end{figure}

As evidenced by Figure~\ref{fig:picp} regarding forecast intervals at all-test and ramp-event samples, the model exceeds a PICP level of 85\% across all forecast periods for the overall test data. In terms of the long-term prediction interval, it is quite close in practice as the values are enclosed within the interval, although slightly larger. Nevertheless, there is a discernible increase in the proportion of samples that fall outside the prediction interval, as soon as the samples are ramp-event based, which would make it ramp-event PICP around 46-47\% in other words.

\begin{figure}[htbp]
    \centering
    \includegraphics[width=0.80\textwidth]{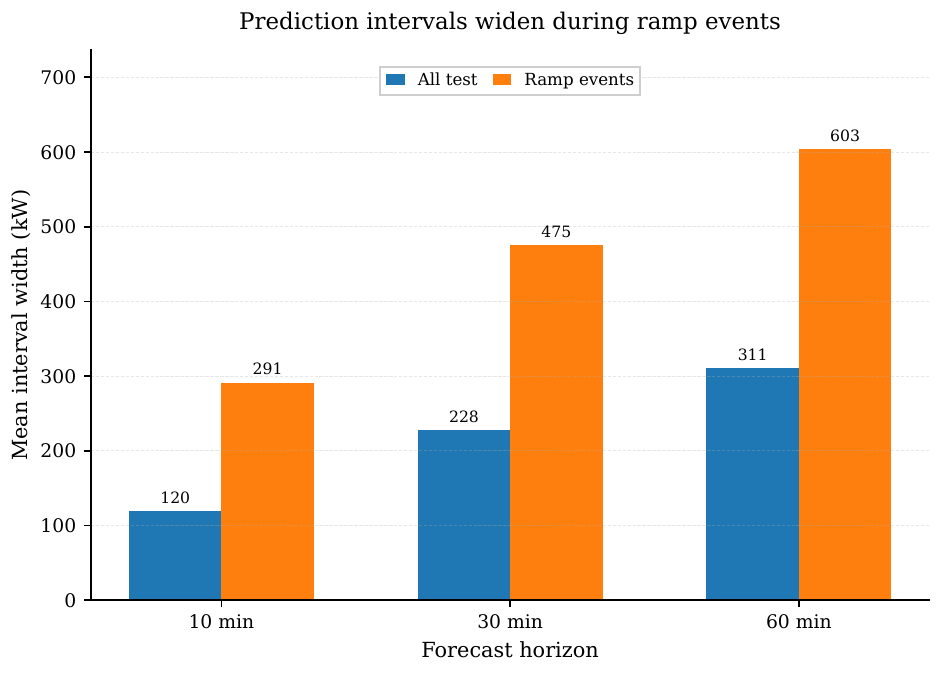}
    \caption{Mean prediction interval width for all-test and ramp-event samples.}
    \label{fig:mpiw}
\end{figure}

In Fig.~\ref{fig:mpiw}, you can view the calculated mean widths for the prediction interval. We note the more expanded nature of the the prediction intervals on the lagged LOAD samples as opposed to the samples from the full set. The values of the computed mean values of the prediction errors for the ramp-event observation is larger than the mean values for the full set only by 142.74\%, 109.39\%, and 94.09\% at the forecast horizons of 10, 30, and 60 minutes, respectively. However, it is noted that in spite of this ever increasing level of expansion, the corresponding ramp-event coverages still fall below the expected 80\% level.

The uncertainty outcomes are outlined in Table \ref{tab:uncertainty_summary}. The results indicate that the quantile model is well aligned with the overall test distribution even though it underestimates the ramp energy event levels. This suggests that ramp events lie at a shift in thresholds where more complex drivers of uncertainty come into play.

\begin{table}[htbp]
\centering
\caption{Uncertainty evaluation for Rich LLMContext quantile XGBoost.}
\label{tab:uncertainty_summary}
\footnotesize
\setlength{\tabcolsep}{3pt}
\renewcommand{\arraystretch}{1.05}
\begin{tabular}{lrrrrrr}
\toprule
Horizon & All PICP & Ramp PICP & Cov. drop & All MPIW & Ramp MPIW & Width inc. \\
\midrule
10 min & 87.74 & 46.80 & 40.94 & 120.88 & 293.41 & 142.74\% \\
30 min & 87.07 & 45.89 & 41.18 & 228.75 & 478.98 & 109.39\% \\
60 min & 85.19 & 47.12 & 38.07 & 321.44 & 623.88 & 94.09\% \\
\bottomrule
\end{tabular}

\vspace{1mm}
\begin{flushleft}
\footnotesize
Note: PICP denotes prediction interval coverage probability; MPIW denotes mean prediction interval width; Cov. drop denotes coverage drop; Width inc. denotes ramp width increase.
\end{flushleft}
\end{table}

These results have a direct impact on operational issues. Testing uncertainty models only against test set endpoints may suggest an optimistic view about the model performance. However experts confirm that when ramps are available, wide intervals can not compensate for the prediction bias. It is, thus, clear that ramps are not only hard to forecast but they are also not having sharper uncertainty measures, this not being due to optimal investment analysis. For grid operators, this under-coverage may result in improper estimation of reserves or lack of speed in adjusting to power changes when necessary.


\section{Discussion}

The recent observations made in this research suggest that the semantical context extracted from LLM is effective, however, its degree in terms of model family, time horizon for the forecasts and evaluation settings can vary. In the out-of-context removal phase Rich LLMContext-XGBoost was best at estimating ramp-event RMSE at 10 and 30 minutes horizons whereas text flicked type of TF-IDF TextContext-XGBoost at 60 minutes. This means that a rich abnormal context has benefits but it is not the case that it is better than all various symbolic or text-based representations.

A more robust evaluation of the role of semantic attributions comes from comparing configurations, with and without declarative content. Enriched in LLM features there was a significant improvement in performance for most of the classification techniques in particular at 30 and 60 minutes. However, most of the gains are not distorted even during resampling and the persona of the ramp-event can be varied because advanced confidence interval and ramp-threshold sensitivity analysis are availed to the reader. Nonetheless, the improvements are in general small so the LLM factor is better addressed as an enhancing class mixture instead of a supplanting factor for limited regression models.

The third section where models and strategies were compared confirms the competitive performance of rich-LLM ensemble, although there are still some other deep sequence models to work on. For the top rich-LLM ensemble the lowest ramp-event RMSE is recorded for both, 10 minutes with GRU, at 30 and 60 minutes with the algorithm performing the best . This demonstrates that there are distinct kinds of knowable information in the temporal system from state changes and the semantic operation context. Moreover, the external validation is relevant to the generalising concept of the socio-technical forecast, but SDWPF provides the most information particularly on ramp-event.

Lastly, the uncertainty distributions reveal that even though the point forecasts have improved, predicting ramp events remains problematic. For example with regard to the prediction intervals, the full coverage is obtained for the full test data but it is not the case for the ramp event samples provided causing the prediction intervals to seriously underperform. This explicitly involves deviation from equilibrium as prediction intervals are based on scales while ramp states have abrupt changes which tend to require special treatment in terms of handling the uncertainties connected with them.

\section{Conclusion}
\label{sec:conclusion}

In this work, a hybrid semantic-context-enhanced scheme for solving the problem of wind power ramp event forecasting was proposed. Instead of using an LLM, as a prediction model, operating modes of wind turbine were represented as a natural language description of the states and embedded as dense representations. These representations were used in combination with the numerical, the lagged, the rolling, the canonical, and the text features in the forms of features over the SDWPF database at forecasts issued for 10-, 30-, and 60- minutes ahead.

Generally, the LLM-derived semantic context tends to add some value but the extent may depend on the model in consideration. In about a tenth of the cases, LLMModels-XGBoost with the richest context ablation domination achieved the best results within 10 and 30 minutes, however, the results of no-LLM compared against rich-LLM approaches were all positive. The same case showed the marginalization of standard hybrid combinations at 60 min and at 90 min. Also bootstrap confidence intervals, a ramp-threshold sensitivity analysis reassure the doubts that many of the reported gains are stable and not merely concerns of one ramp size or one sample test more than others.

A more recent study revealed that the existing rich-LLM ensemble models still failed to weaken strong conventional and deep sequence models. The rich-LLM ensemble with the best RMSE attained prediction results that are more accurate than GRU or any other such models for a 10-minute ramp-event, whereas for the 1-hour ramps, the best single task model was GRU. To summarize, The motivation of this study is not a pure LLM or its variants forecasting a model but the concept of the operating context which can enhance the existing forecasts in general.

In a combined validation including data from Kaggle SCADA and NREL wind databases it was indicated that interfered models which included even large language models reduced the root mean square error across the horizons evaluated albeit that it was mainly in the ramps forecasted by the LLM-test database. Examining the variation in the performance, it can be observed that the uncertainty about the estimates that the authors derived was very appropriate on these samples which confirms certain validity but the authors did not calibrated the uncertainty for ramp events.


\section*{Declaration of Competing Interest}
The authors report no financial, professional, or personal interests that could be perceived as having influenced the research presented in this manuscript.

\section*{Acknowledgements}
The authors gratefully acknowledge the support and contributions that facilitated the completion of this research.

\section*{Data Availability}
The datasets associated with this study may be obtained from the corresponding author upon a reasonable request.



\bibliographystyle{unsrtnat}
\bibliography{references}

\end{document}